\RequirePackage{fix-cm}
\documentclass[twocolumn]{svjour3}          
\smartqed  
\usepackage{graphicx}
\usepackage{subfig}
\usepackage{natbib}
\usepackage{amsmath}

\usepackage{lineno,hyperref}

\usepackage{multirow}
\usepackage{dblfloatfix}
\usepackage{tikz}
\usetikzlibrary{positioning}

\usepackage{algorithm}
\usepackage{algpseudocode}
\usepackage{color}
\usepackage{array}
\usepackage{multirow}
\usepackage{booktabs}
\usepackage{calc}
\usepackage{lipsum}
\usepackage{amssymb}
\usepackage{wrapfig}

\begin{document}

\title{Weakly-Supervised Road Affordances Inference and Learning in Scenes without Traffic Signs
}


\author{Huifang Ma \and Yue Wang \and Rong Xiong \and Sarath Kodagoda \and Qianhui Luo
}


\institute{Huifang Ma, Yue Wang, Qianhui Luo and Rong Xiong \at
	State Key Laboratory of Industrial Control Technology and Institute of Cyber-Systems and Control, Zhejiang University, Zhejiang, China \\
	\email{rxiong@zju.edu.cn}           
	\and
	Sarath Kodagoda \at
	Centre for Autonomous Systems, The University of Technology, Sydney, Australia
}

\date{Received: date / Accepted: date}

\maketitle

\begin{abstract}
Road attributes understanding is extensively researched to support vehicle's action for autonomous driving, whereas current works mainly focus on urban road nets and rely much on traffic signs. This paper generalizes the same issue to the scenes with little or without traffic signs, such as campuses and residential areas. These scenes face much more individually diverse appearances while few annotated datasets. To explore these challenges, a weakly-supervised framework is proposed to infer and learn road affordances without manual annotation, which includes three attributes of drivable direction, driving attention center and remaining distance. The method consists of two steps: \textit{affordances inference from trajectory} and \textit{learning from partially labeled data}. The first step analyzes vehicle trajectories to get partial affordances annotation on image, and the second step implements a weakly-supervised network to learn partial annotation and predict complete road affordances while testing. Real-world datasets are collected to validate the proposed method which achieves 88.2\%/80.9\% accuracy on direction-level and 74.3\% /66.7\% accuracy on image-level in familiar and unfamiliar scenes respectively.
\end{abstract}

\section{Introduction}
\label{Introduction}
Road attributes understanding is widely researched for autonomous driving, which relates to various tasks such as road area segmentation, traffic sign detection, intersection recognition, et.al. For the urban road nets where comprehensive traffic signs and driving rules are provided, there is an emerging topic to learn \textit{affordances}, attributes of the environment which limit the space of allowed actions\citep{sauer2018conditional}. Under specified driving situations, the definition of affordances involves both road attributes as well as attributes of other traffic participants relating to vehicle action, e.g. distance to surrounding lanes\citep{chen2015deepdriving}, distance to neighboring cars\citep{chen2015deepdriving}, and existence of red traffic light\citep{sauer2018conditional}. Compared with the mediate sub-components for recognizing driving-relevant objects, the affordances are more relevant to driving decision making and can be efficiently trained with advanced neural networks due to the low-dimensional representation.

However, there are massive scenes where roads are not in standard conditions and traffic signs are absent or far from satisfactory, such as in campuses and residential areas. In these scenarios, there are few restrictions on vehicles' positions and movings. Thus related research basically fall in the conventional map matching framework\citep{qin2018relocalization}\citep{ding2018laser}\citep{tang2018topological}. The road attributes understanding is limited to traversable area segmentation\citep{barnes2017find}\citep{Tang2018From}\citep{wang2019ral} and intersection recognition \citep{bhatt2017have}\citep{kumaar2018juncnet}\citep{kumar2018towards}. And the dynamic traffic participants are addressed as obstacle avoidance. To make a step further based on the environment characteristics, we propose to introduce the concept of affordances to these scenes as \textit{road affordances}, attributes of road structure which support vehicle actions. Without tips from traffic signs, it has long been challenging to learn road structure directly from visual appearance due to environment complexity. There are large variations in viewpoint, illumination, and appearance across roads, while not enough annotated datasets. Nevertheless, the avoidance of obstacles can be achieved with mature vision modules or additional range finder sensors.

In regard to the challenges of lacking dataset and prior definition, we rethink how the initial semantics of road structure is established by human, and notice that to summarize past experience and discover their rules has been an inherent ability for human to gain knowledge. Therefore, a non-parametric method is developed in this paper to firstly infer road affordances from driving behaviors and then learn from the inferred affordances for end-to-end prediction, which does not need any manual annotation. The method has achieved a description of road affordances including three attributes, which are \textit{drivable direction}, \textit{driving attention center} and \textit{remaining distance to attention center}. We consider the three attributes are valid to guide vehicle action, as the information of \textit{turn right after 8m in next intersection} is constantly used by humans to navigate themselves. The \textit{turn right} command corresponds to a recognition of drivable directions, the \textit{8m} is related with a rough estimation of distance, and \textit{in next intersection} involves visually localizing a turning area to pay attention to.

The intuition behind affordances inference is representing road structure with vehicle trajectories. In this way, analyses of road attributes can be achieved by analyses of trajectories which are more informative with driving behavior features. We specifically use a HDP-HMM model to segment trajectories based on the feature of angular speed. Different angular speeds show different allowed driving actions, and each indicates a drivable direction. Then the segmented trajectories are projected on concurrently collected visual observations to infer road affordances. However, the inferred road affordances is incomplete in this step, since the annotated trajectory only covers one possible driving action that could be taken at each place while there are multiple action choices when approaching intersections and T-junctions. And to collect trajectories data for all the choices of driving action needs substantial preparatory work.

\begin{figure*}[!ht]
	\centering
	\includegraphics[width=0.9\textwidth]{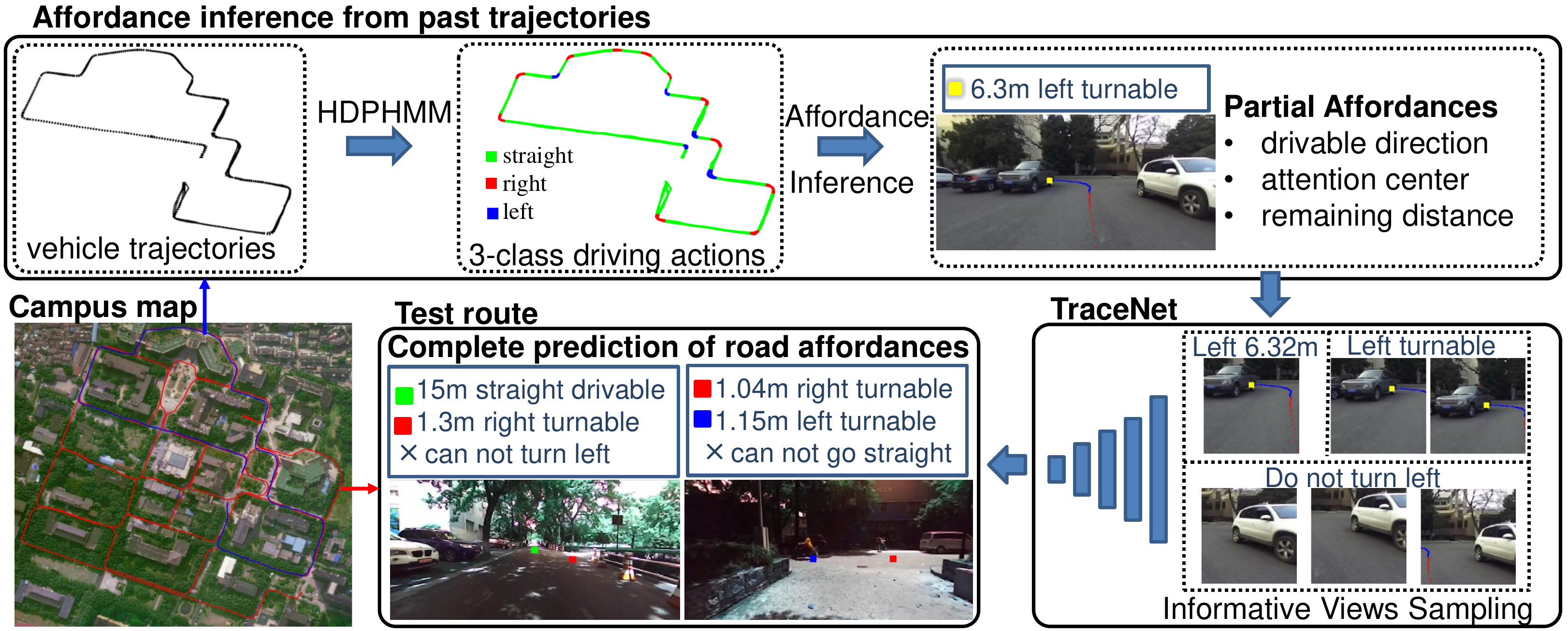}
	\caption{Outline of the proposed method. The first step is partial affordance inference along demonstration route, which analyzes affordances by projected semantic poses. The second step is learning from incompletely labeled data in TraceNet, which enables a complete prediction to road affordances.}
	\label{outline}       
\end{figure*}

Then the following step is to learn from the partially annotated data and train an end-to-end model directly from image input to complete road affordances even in previously unseen scenarios. We address the learning task with a weakly supervised multi-task network called ``TraceNet''. To specifically handle the annotation deficiency, TraceNet takes local image regions as input, which are sampled in different ways based on the position of annotated attention center. The local regions contain different affordance information which may be incomplete for all the three attributes while contribute more concentrated partial supervision to certain sub-tasks. Then the partly unknown information of local regions is explicitly modeled in the loss function to enable co-training. While testing, the network is performed on the original image size in a sliding window manner to predict complete road affordances. The outline of the two steps are provided in Fig. \ref{outline}.

The proposed method is validated on the real world data sets collected in our school: YQ21\citep{tang2018topological} and YQ-South\citep{tang2018topological}. Experimental results indicate the effectiveness of both affordances inference from trajectory and affordances learning with TraceNet. The model has achieved 88.2\%/80.9\% accuracy on direction-level and 74.3\% /66.7\% accuracy on image-level for affordances prediction in familiar and unfamiliar scenes respectively. Our main contributions are as follows:
\begin{itemize}
	\item A novel approach is developed for road affordances learning in regard to the scenes without traffic signs. The approach learns directly from visual appearance without manual annotations and manages to predict complete road affordances including drivable directions, driving attention centers, and distances to attention centers. 
	\item An inference model from vehicle trajectories to road structure attributes is established with a HDP-HMM clustering method to analyze driving behaviors. The correlation of driving behaviors with road structures is innovatively utilized to generate visual annotation.
	\item A multi-task network called ``TraceNet'' is implemented to learn from partially labeled data. The network explicitly models the unknown information in the loss function and achieves co-training with complementary partial supervisions.
\end{itemize}
The rest of this paper is organized as follows: Related works are reviewed in Section \ref{Related works}; The proposed method is illustrated in Section\ref{method}, including affordances inference model from trajectory in Section \ref{method1} and design of TraceNet in Section \ref{method2}. Experimental results and discussions are presented in Section \ref{experimental result}, and conclusions are drawn in Section \ref{conclusion}.

\section{Related works}\label{Related works}
\paragraph{Urban Road Nets.} Large amount of works have been carried out on visual perception with urban road net. Those literature can be categorized into two groups. 
The first devotes to recognition of road-relevant objects, including lane detection\citep{lee2017vpgnet}, traffic sign recognition\citep{zhu2016traffic}\citep{Ruhhammer2017Automated}, car detection\citep{li2019stereo}\citep{li2018stereo}, as well as overall scene understanding involving multiple recognition and segmentation tasks\citep{Xun2017Semantic}. The recognition results are then combined into a consistent world representation of the vehicle's immediate surrounding for following decision making.
The second group tries to pursue a higher level of direct understanding while still relies on the structured traffic infrastructures. Geiger et,al \citep{geiger20143d} aimed to estimate intersection topology and geometry as well as to localize other traffic signs using a set of hand-crafted image features. Seff et,al\citep{seff2016learning} leveraged the standard navigation maps and corresponding street view images to learn multiple road layout attributes in vision images. The concept of affordances is first proposed by Gibson\citep{gibson2014ecological} in the field of psychology and has been applied to autonomous driving task in the work of Chen et,al\citep{chen2015deepdriving}. They proposed to predict driving affordances in the highway situations, which includes vehicle pose, distances to the lanes, and distance to other road participators. Al-Qizwini et al. \citep{al2017deep} then improved the work in \citep{chen2015deepdriving} by analyzing different CNNs for the mapping from image to indicators, and find GoogleNet and VGG16 to perform best on this task. And later the work in \citep{sauer2018conditional} generalized the affordances learning in \citep{chen2015deepdriving} to more complex driving scenarios and involves more affordances from traffic signs. Recently, the work in \citep{yi2019how} considered driving intentions from other traffic participants to constrain vehicle actions, which is also a kind of environment affordance. Our work is inspired by the second group of direct understanding in urban road nets, while we have pay special attention to the scenes without traffic signs.

\paragraph{Scenes without traffic signs.} There are massive scenes that generally lack of traffic signs and related common features. Thus, research on intersection recognition have been developed using both 3D data\citep{zhu20123d}\citep{zhang20153d} and image data\citep{bhatt2017have}\citep{kumaar2018juncnet}\citep{koji2019deep}\citep{kumar2018towards}. Since human drivers can use experience and general knowledge to handle such complex scenes, there are works utilizing prior information for autonomous perception. Levinkov et,al\citep{Levinkov2013Sequential} proposed to annotate diverse road scenes based on a Bayesian update model with the prior knowledge from structured traffic scenes. Guru et,al\citep{Gur2017Learn} proposed to learn past experience to predict the likely detection failure of a robot in the known workspace, and enable robot to hand over tasks to human when the perception system tends to have poor performance. The weakly-supervised drivable area segmentation method proposed in \citep{barnes2017find} assumes the traversal areas are exactly where vehicle has visited earlier and can be projected to images as annotation. However, this system can only accommodate one trajectory of the data collection route, which limits the driving span of a road scene. In later developments, Tang et,al\citep{Tang2018From} extended the trajectory projection from one to many, which broadens the traversable area. These works without dependence on traffic signs are basically fall in the first group of research in urban rod nets, while we denote to the second group of visual representation which can be used to facilitate more intelligent navigation strategies.

\paragraph{Learning from partially labeled data.} This part gives a brief review of works which have a similar annotation setting as ours for network training. The problem of training classifiers with positive and unlabeled samples is called PU classification (trajectory annotation only provides positive labels and other possibilities are unlabeled). There are some studies attempted to address this problem in a binary setting. Elkan et,al \citep{Elkan2008Learning} constructed a probabilistic model using observable samples to design a classifier. Plessis et,al\citep{Plessis2014Analysis} revealed that PU classification can be cast as a cost-sensitive learning, which changed the penalty weights of each candidate class. Other works have attempted to address label incompleteness in multi-label learning as label deficits. Bucak et,al\citep{Bucak2011Multi} tried to eliminate the influence of label deficits by adding a regularization term to rank loss, which forces the differences between scores of positive and negative labels to be group sparse. Kong et,al\citep{Kong2014Large} extended the work in \citep{Elkan2008Learning} to a multi-label setting by considering the label dependencies. Subsequently, a conditional restricted Boltzman machine was used in \citep{Li2015} to handle the deficiency on labels. These works mainly focused on improvement of rank loss to achieve higher classification accuracies. The loss function in this paper is inspired by these works, while addresses multiple sub-tasks for annotation deficiency rather than the only classification.
\section{Method}\label{method}
The outline of the proposed framework is shown in Fig. \ref{outline}. In the first step, vehicle trajectories are collected along the demonstration route shown with blue lines on the campus map. The sequential angular speeds of trajectories are then fed into a HDP-HMM model for road interval segmentation, which can generate a 3-class result of driving actions including \{\textit{straight-drivable, right-drivable, left-drivable}\}. Then the segmented trajectories are projected on corresponding vision observations to infer partial road affordances in relation to the assigned driving action. The second step learns from the partial affordances with a specifically designed network called ``TraceNet'', which performs informative view sampling to incorporate different partial supervisions and explicitly models the unknown label in the loss function. TraceNet can predict complete road affordances in a longer test route shown with the red lines on the campus map. The following sections provide the details of each step.
\subsection{Affordances inference from trajectories}\label{method1}
\begin{figure}[!h]
	\includegraphics[width=0.48\textwidth]{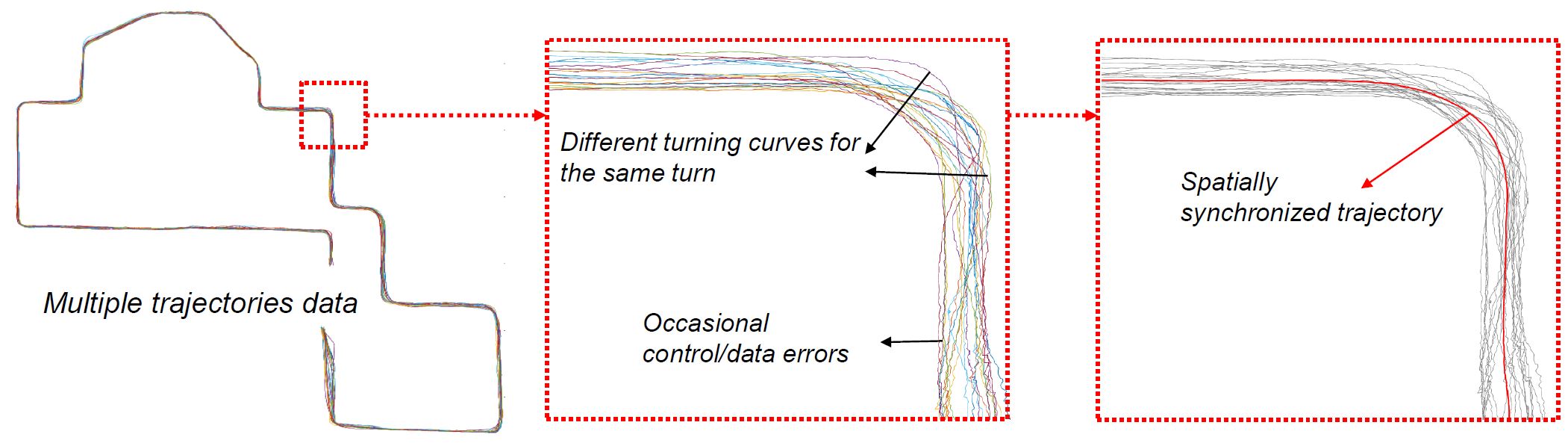}
	\caption{Individual differences in multiple trajectories.}
	\label{trajectories}
\end{figure}

Since there are individual variations on vehicle control and unavoidable data errors, a single trajectory may not be representative for road structure. Thus, multiple trajectories of different running are collected for road affordance inference, as shown in Fig .\ref{trajectories}. 

The trajectories from different runnings are spatially synchronized to calculate statistical angular speeds for driving behavior analysis\citep{ma2018time}, which eliminates the impacts of single trajectory and better represents the allowed driving action related to road structure. The vehicle poses on trajectories are obtained with traditional localization algorithm\citep{tang2018topological} and the angular speeds are estimated with recorded timestamps. 

\subsubsection{Trajectory segmentation with HDP-HMM}
\begin{figure}[!h]
	\centering
	\includegraphics[width=0.49\textwidth]{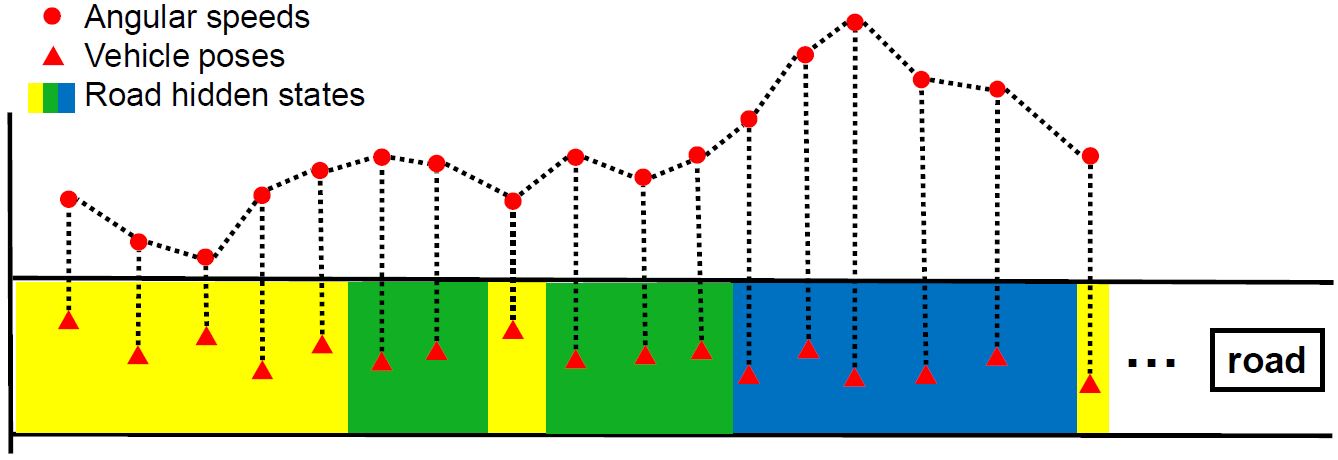}
	\caption{Hidden Markov Model for a road section. Different colors on the road indicate different allowed driving actions.}
	\label{hmm}
\end{figure}

The relation between angular speeds and physical road locations is modeled with a HMM(Hidden Markov Model), as shown in Fig. \ref{hmm}. The horizontal axis shows the stretch of traversed road. Different colors on the road imply different hidden states of allowed driving actions which are currently unknown. Red triangles and red dots denote vehicle poses and angular speeds respectively.

In general, it may be adequate to directly define a 3-class road states of \{\textit{straight-drivable, right-drivable, left-drivable}\} according to human common sense for most traffic situations. We expect to explore more possibilities on driving behaviors in a knowledge induction manner. Therefore, we do not assume a prior number of road states for classification and instead use a sticky HDP-HMM(Hierarchical Dirichlet Process) model\citep{Fox2008An} to sequentially clustering angular speeds to several groups. The HDP-HMM model is first proposed in \citep{Yee2006Hierarchical}, which employed a stochastic process of HDP as transition probability in HMM to indicate the unknown number of transition states. Then the sticky HDP-HMM adds an extra parameter in HDP-HMM to bias the process toward self-transition, which is more suitable for systems with state persistence, just as the road intervals in our case. 

The DP(Dirichlet Process) is a measure on measures, which shows a discrete clustering property subject to a continuous base measure. The discrete clustering groups can be regarded as a parameter sampling result from a continuous function, thus different groups have totally different distribution parameters. For HDPs, the base measure is a DP process which is discrete and nonparametric. Therefore, the clustering groups are sampled from a discrete base measure, which necessarily share distribution parameters. Then the similar driving actions can be assigned to discontinuous road intervals during sequential clustering. The mathematical proof is beyond the scope of this paper and more formulas can be inferred in \citep{Yee2006Hierarchical}.

In the experiment, different number of groups are obtained by tuning model parameters, among which the 3-group result and 5-group result are more consistent with human intuition. The 3-group result is the three primitive classes of \{\textit{straight-drivable}, \textit{right-drivable}, \textit{left-drivable}\}. And the 5-group result further distinguishes the turn-in/turn-out areas of \textit{left-drivable} and \textit{right-drivable} respectively. Therefore, human driving patterns can be learned in a non-parametric manner, which associate with road structures. Considering the turn-in/turn-out areas are actually the transition stages among the 3-class result, which are not visually distinctive for scenes without traffic signs. The 3-class result is used for the following vision affordances annotation.

\subsubsection{Partial affordances annotation}\label{annotation}
\begin{figure}[!h]
	\centering
	\subfloat[turning class]{\includegraphics[width=0.46\textwidth]{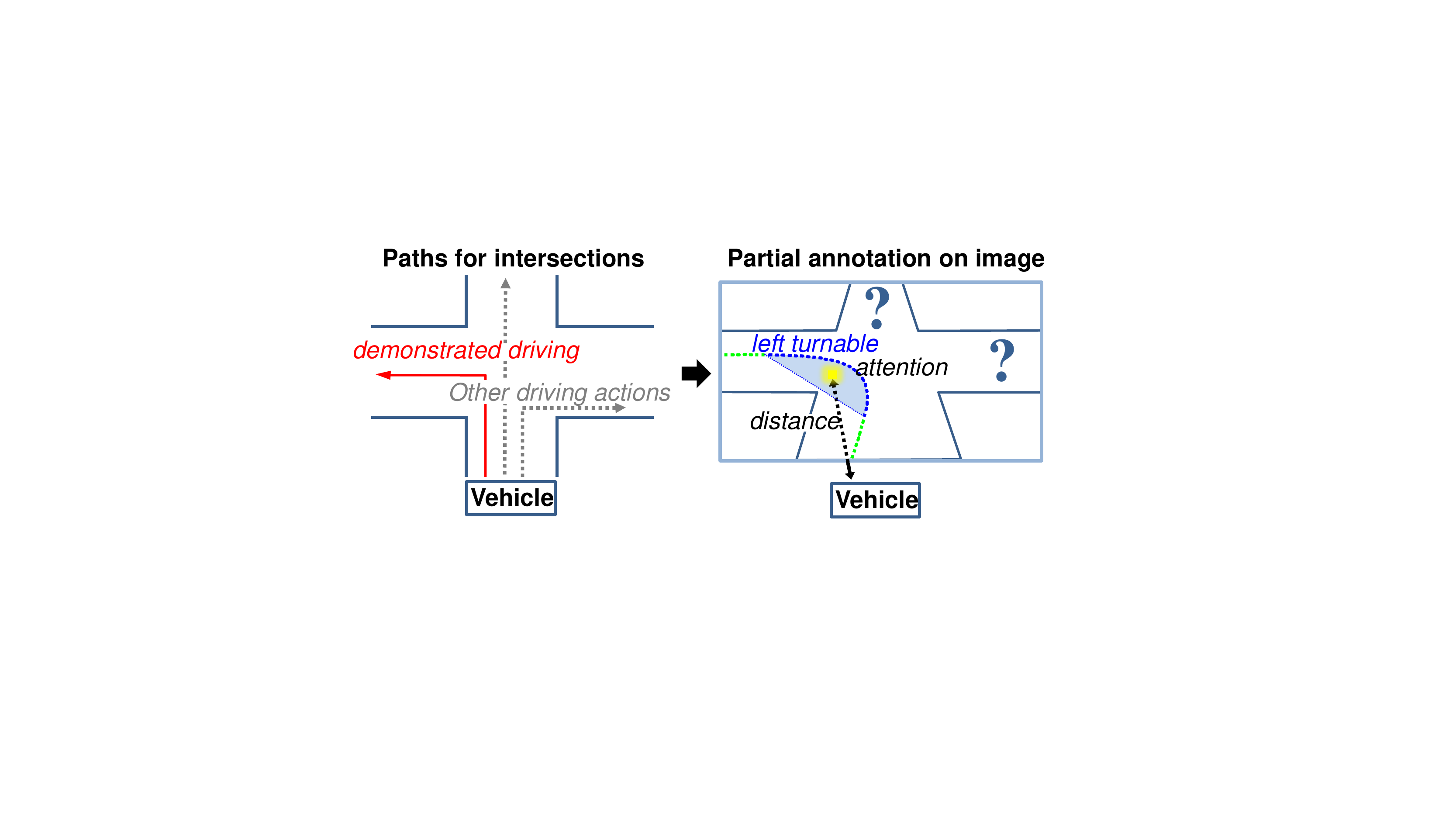}}
	\\
	\subfloat[straight class ]{\includegraphics[width=0.46\textwidth]{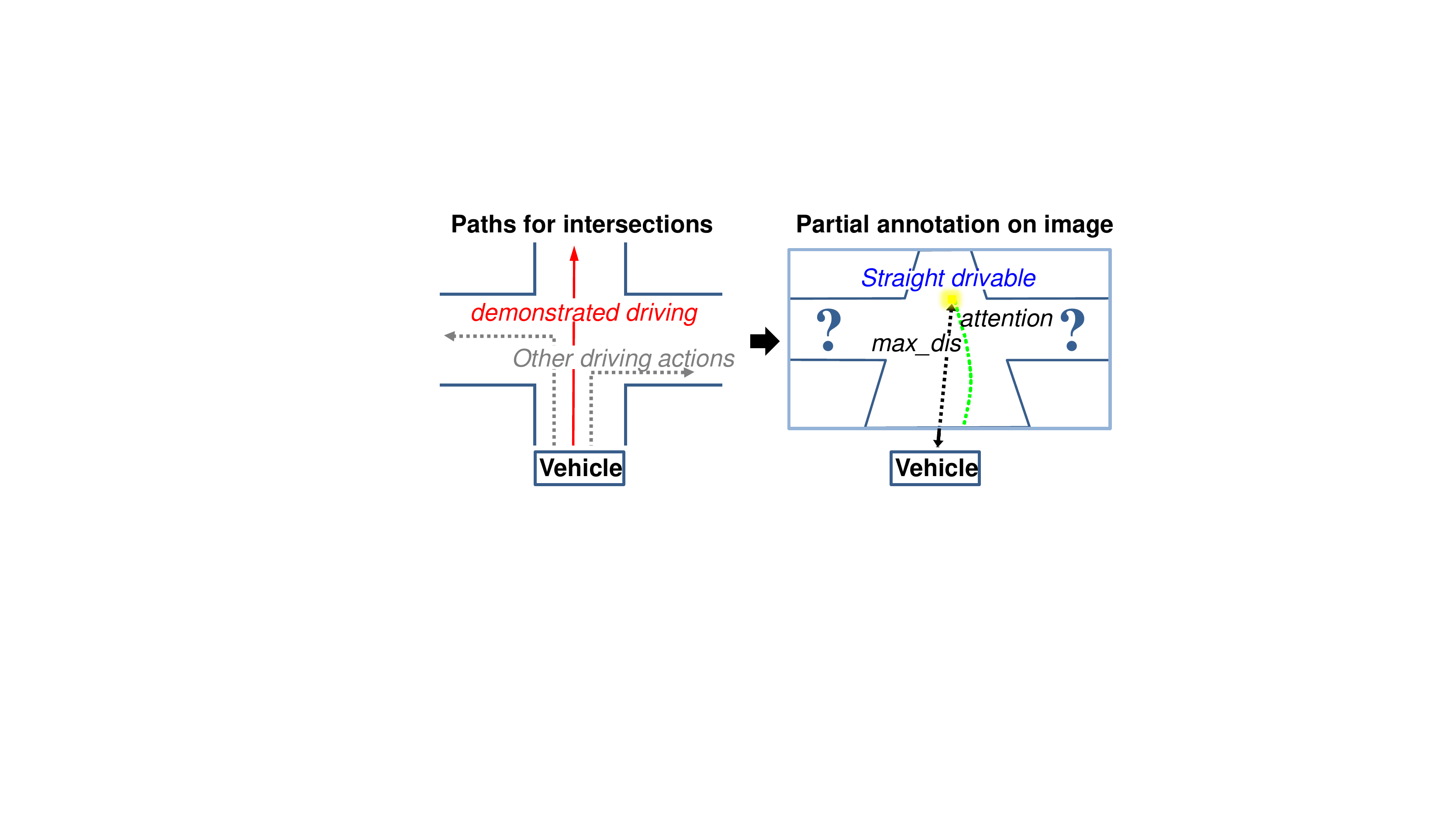}}
	\caption{Partial affordances inferred from trajectory. The colored dots on images are vehicle poses. Green color shows the straight driving poses and blue color shows the left turning poses. Yellow squares are the driving attention centers.}
	\label{partial annotation}
\end{figure}

When trajectories are segmented and assigned with driving actions, the vehicle poses within a distance of 15m to the current location are projected on concurrent vision observation. And road affordances relating to the demonstrated driving action can be analyzed accordingly as illustrated in Fig. \ref{partial annotation}.

Fig. \ref{partial annotation}(a) and Fig. \ref{partial annotation}(b) show the cases of turning classes and straight class respectively. The visual road affordances of drivable direction, driving attention center and remaining distance are defined as follows:

\textbf{Drivable direction} is annotated by whether there are \textit{right}/\textit{left} turning poses projected within the maximum prediction distance, otherwise it is the \textit{straight-drivable} class. \textbf{Attention center} is defined as the center of projected turning poses for \textit{right}/\textit{left} classes, as shown with the yellow square in Fig.\ref{partial annotation}(a). For the \textit{straight} class, it is defined as projection of the pose meeting maximum distance constraint, as shown in Fig. \ref{partial annotation}(b). \textbf{Distance} is measured as the straight line distance from current location to the center pose used for attention center projection. Since there can be different routes leading to the attention center, the straight line distance shows a better definite relation to attention center than the piecewise distance of pose trajectory, which is adopted in our method. For the turning poses, the distance is defined as 0.

As implied in each subfigure, the demonstrated driving only reflects one allowed driving action at each place, while no information for the other possible actions. It leads to incomplete understanding to road structures and can only achieve partial affordances annotation on images. A complete demonstration of driving actions needs substantial work for trajectory and sensory data collection, i.e., it need 12 times of driving for each intersection (4 entries $\times$ 3 exits), let alone other small footpaths and T-junctions. Therefore, we devote to the learning-based manner in the next step to achieve prediction on complete road affordances.
\begin{figure*}[!ht]
	\centering
	\includegraphics[width=0.9\textwidth]{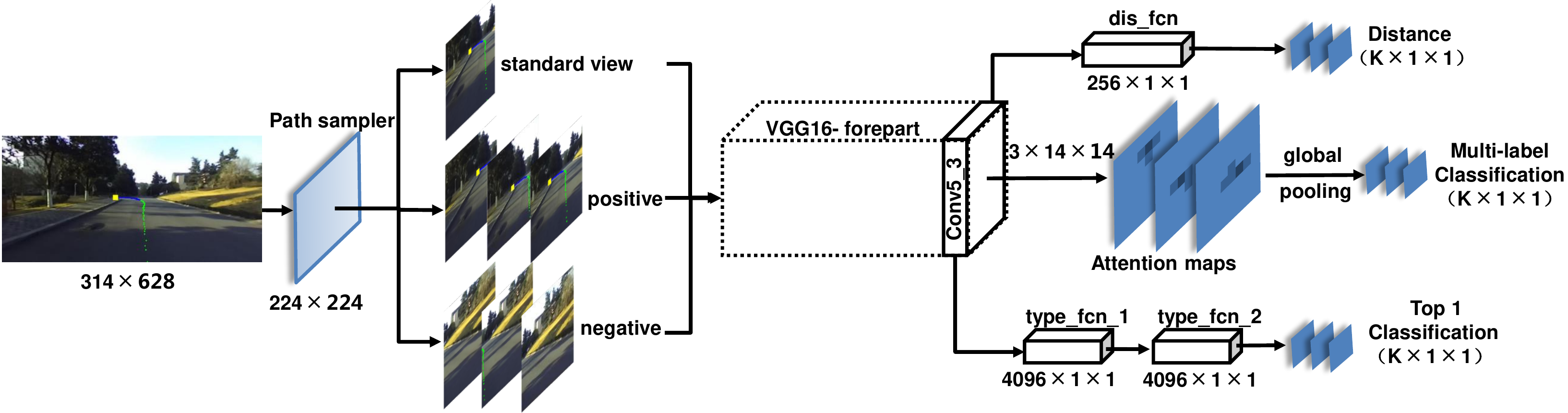}	
	\caption{Network structure of TraceNet while training.}
	\label{train-net}
\end{figure*}
\subsection{Learning from Partially Labeled data}\label{method2}
This section illustrates the proposed TraceNet which learns from partially labeled data to predict complete road affordances. To clarify the problem, the partial annotation can be stated as follows: 
1) Assigned label relates to a local image region.
2) Assigned labels are definitely positive.
3) Absent labels are not necessarily negative.
Statement 1. implies the task belongs to object localization, which needs to localize an attention center for each driving action. Statement 2. and 3. have been studied as positive and unlabeled (PU) classification problem in the binary case. As the affordances allow to contain more than one class, our learning task is multi-label localization under PU labels.

For multi-label prediction, PU problem means each label actually refers to a related local region instead of the complete image and the information of other regions is unknown. Moreover, each positive label needs to regress a physical distance in our case. In order to make the learning tractable, we make two assumptions for this specific case:
1) Samples have non-repetitive labels.
2) Samples have non-overlapping local regions for existing classes.
The two assumptions are usually satisfied in practice: vehicles usually can not see two intersections within a restricted point of view and different road branches are surely separated. Under these assumptions, different classes are visually separated. And TraceNet can process on a smaller size of local regions which only contain one possible class and are more focus on the limited annotation. The network structure is shown in Fig. \ref{train-net}.

The original image is firstly processed by a controlled view sampling layer which extracts three types of local views including standard view, positive views and negative views. These local views contain information on different aspects of road affordances, which will be more specifically illustrated in Section \ref{viewsamples}. Then the local views go through a VGG16-forepart block\citep{simonyan2014very} for common feature extraction and the multi-task branches for affordances prediction. There are three multi-task branches used for network training:

1. Multi-label classification branch predicts the existence probability of each drivable class along with the location of attention center via global max-pooling layer. The attention center is at a resolution of $16 \times 16$ pixels.

2. Distance regression branch is trained to estimate a remaining distance to attention center for each existing class. 

3. Top1 classification is a traditional multi-class classification branch, which helps to activate the networking training.

While testing, the network is performed in a sliding window manner on the original image to localize road structures. Following gives specific illustrations of the weakly-supervised loss function, local view sampling, training initialization as well as inference while testing.

\subsubsection{Weakly-supervised loss function}
The multi-task loss function is a weighted sum of the three branches:
\begin{equation}
	Loss = w_1L_{multi\_label} + w_2L_{distance} +w_3L_{top1}
\end{equation}
where $w_1$,  $w_2$ and  $w_3$ are weight terms, which are set reversely to the observed magnitude of three loss values. Since each subtask can face incomplete supervision from the local views, the loss functions have all explicitly modeled the unknown labels in their formulations.

\paragraph{Multi-label Classification Loss.} The aim of this branch is to output a single score indicating the positive possibility for each of the driving class as well as its attention center. The training of attention center does not rely on the pixel-level annotation from trajectory, as it may be less accurate to represent a distinctive region. This branch has referred to the work in \citep{oquab2015object}, which aggregates the $n\times n\times K$ matrix of output scores for $n\times n$ different positions of the input window into a single $1\times 1 \times K$ vector using a global max-pooling operation, where K is the number of classes and equals to 3 in our problem. The max-pooling operation effectively searches for the best-scoring candidate regions within the input image, which is crucial for attention center learning. In addition, due to the max-pooling operation, output of the network becomes independent with the size of input image, which can be used for the original image inference in the test phase.

The loss function of Multi-label classification is defined as:
\begin{equation}
	\begin{aligned}
		& L_{multi\_label} = \sum_k log(1+e^{-y_kf_k(x)}) \\
		& y_k\in\{1,0,-1\} 
	\end{aligned}
\end{equation}
where $y_k$ indicates the supervision information of class $k$, $y_k=1$ means positive sample, $y_k=-1$ means negative sample, and $y_k=0$ means no information for class $k$.

\paragraph{Distance Loss.} This branch estimates a remaining distance to the driving attention center of each class, with the ground truth provided by the straight line distance from the current pose to the pose regarded as attention center. The loss function uses L1 distance for regression:
\begin{equation}
	\begin{aligned}
		&L_{distance} = \sum_k sign(y_k)\cdot |f_k(x)-y_k| \\
		&y_k\in[0,max\_dis]
	\end{aligned}
\end{equation}
where $y_k$ is a continuous value ranging from 0 to the maximum prediction distance. $y_k>0$ means a valid distance supervision for a positive straight class, $y_k=0$ means a valid supervision for turning class, and $y_k<0$ means no supervision for distance.

\paragraph{Top1 Loss.} At the beginning of training, each iteration of partial supervision tunes only a small part of parameters, which can not provide sufficient gradient for the multi-label branch to learn well. In this case, we have added a traditional multi-class classification branch to initialize the training of network, which forces parameters to generate at least one strong response to the positive samples and thus distinguish different classes. Top1 loss takes the form of softmax and the supervision comes from standard/positive views which provide definite class information. The improved softmax loss function that explicitly considers the unknown label is denoted as follows:
\begin{equation}
	\begin{aligned}
		&L_{top1} = -\sum_k sign(y_k)\cdot (y_k + 1) \cdot log(\frac{e^{f_k(x)}}{\sum_k e^{f_k(x)}}) \\
		&y_k\in\{1,0,-1\} \\
		&\textbf{s.t.} \quad \sum_k|y_k| = 0, \quad if \prod_k sign(y_k) = 0
	\end{aligned}
\end{equation}
where $y_k$ indicates the supervision information of class $k$. For positive sample with class $i$, only $y_{i}$ equals to 1 and other values in the label set equal to -1, namely $\{y_{k} = -1| k\in K,k\neq i\}$. 
For negative samples which only provide partial negative information and do not contain information of the unassigned classes, all the values in the label set equal to 0, namely $\{y_{k} = 0| k\in K\}$, thus the loss equals to 0.

\subsubsection{Informative View Sampling}\label{viewsamples}
\begin{figure*}[!h]
	\centering
	\subfloat[Standard view]{\includegraphics[width=0.32\textwidth]{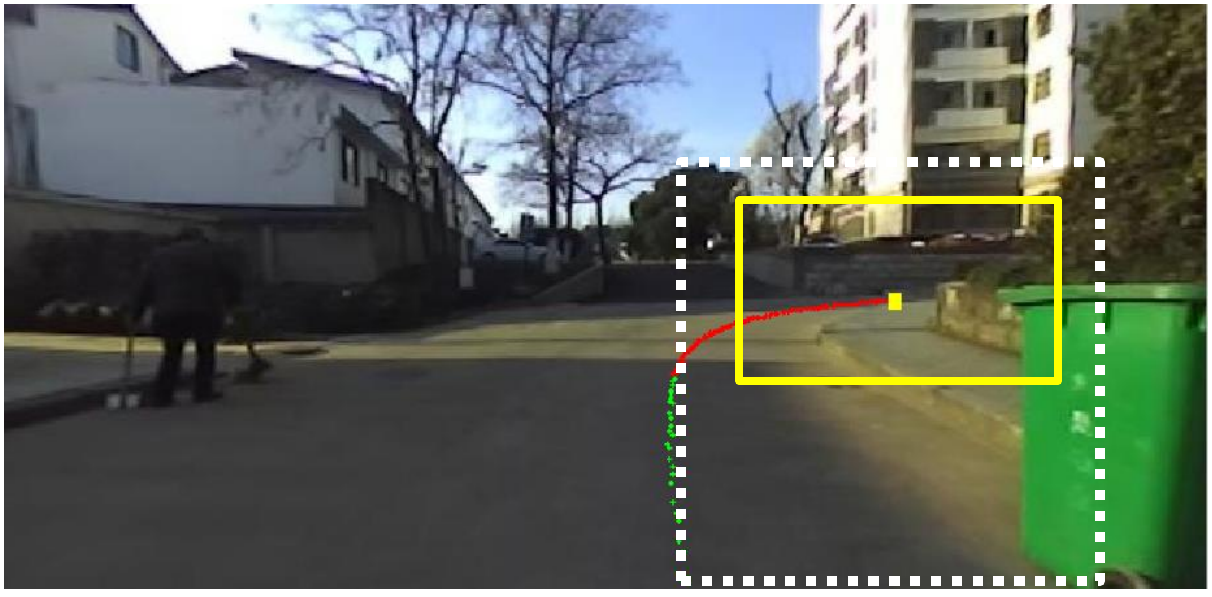}}
	\hspace{0.0001\textwidth}
	\subfloat[Positive view samples]{\includegraphics[width=0.32\textwidth]{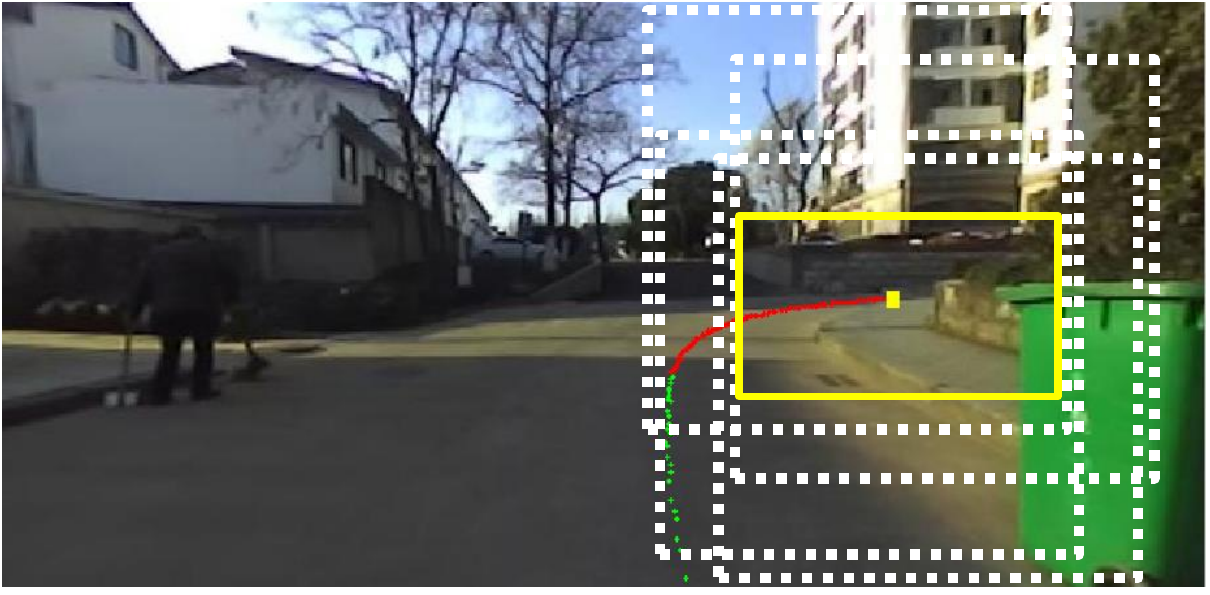}}
	\hspace{0.0001\textwidth}
	\subfloat[Negative view samples ]{\includegraphics[width=0.32\textwidth]{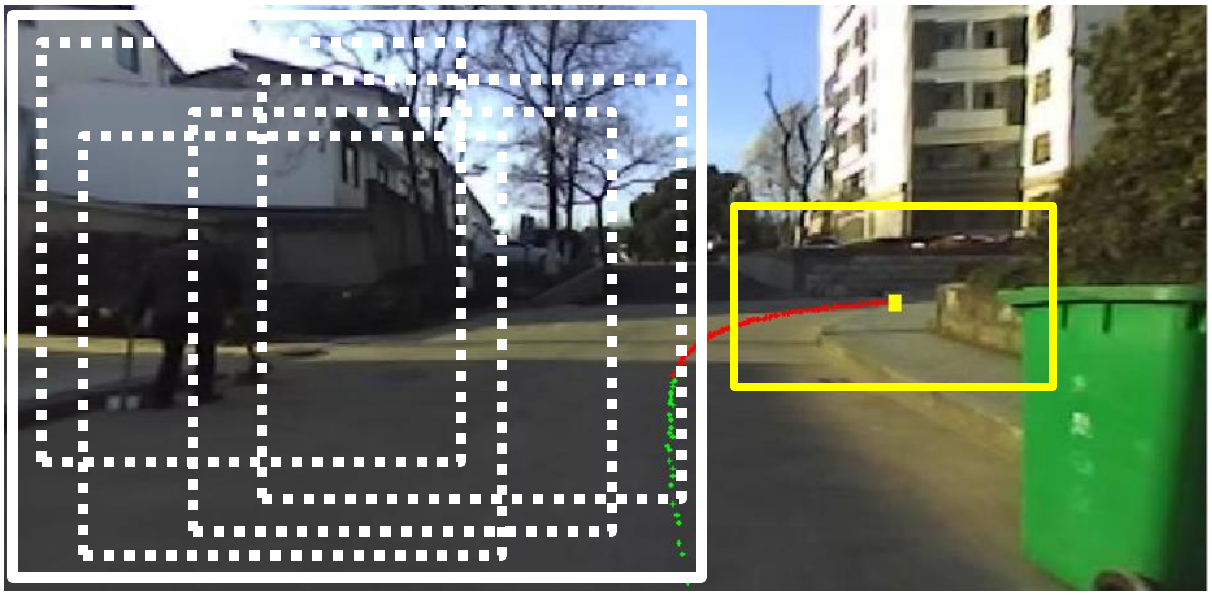}} 
	\caption{Three types of views used for network training. The red and green dots on the road show the projection of segmented trajectory. The yellow dot indicates the point of attention center. Box in yellow is the safe zone that controls the sampling region of positive and negative views, and white dotted boxes are the example samples of each view type.}
	\label{views}	
\end{figure*}

The three types of local views are sampled based on the projected trajectory. Fig. \ref{views} illustrates the sampling process with a \textit{right-drivable} sample.

The local views are extracted based on a safe zone centered on the annotated driving attention center. The safe zone needs to be smaller than the size of local view and is experimentally set as $90 \times 160$ pixels. Different types of views are illustrated as follow:

\paragraph{Standard view} is a fixed view that horizontally centers at the annotated attention center and vertically reaches the bottom of image. It contains the positive information for assigned class and is the only one that can provide supervision for distance regression. Since the camera is at a fixed height relative to the ground, pixels of road area with the same row on the image roughly indicates a similar visual range. The vertical position of attention center on image is thus a direct measure to the remaining distance. 

\paragraph{Positive views} are randomly sampled views that contain the safe zone. These views guarantee the annotated attention center appear on different positions of the views, which helps on the convergence of driving attention center regardless of rotation and transformation. The global max pooling step ensures the network will respond to the most commonly distinctive regions. But, the random offsets on vertical direction have lost the consistent reference to the physical world and can not be used for distance regression.

\paragraph{Negative views} are randomly sampled views that avoid the safe zone. Under the non-overlapping assumption, the region away from the assigned attention center are not belonging to the assigned label. Thus negative views provide the necessary negative information for the current assigned class. Nevertheless, it is still not sure whether it belongs to any of the other absent classes, and can only provide partial negative information to the multi-label classification task.

The local view sampling is implemented in a \textit{Path Sampler} layer and different types of samples are controlled in a roughly equal proportion. 

\subsubsection{Training Initialization}
Since the \textit{straight-drivable} class takes a dominant proportion in the training data, we have donwsampled the straight class to approximately one sixth, which leads to about twice of the samples for turning classes. Downsampling aims to solve the data imbalance, while it still needs to keep a distribution roughly representing the real case, where straight-only roads take a big proportion in most scenarios and such turns should be strictly forbidden. Then for an overall data augmentation, we have randomly mirrored the images and re-marked image labels: the \textit{left}/\textit{right-drivable} image mirrors to the \textit{right}/\textit{left-drivable} image, and the \textit{straight-drivable} label remains unchanged.

The VGG16-forepart block is initialized with the pre-trained parameters in \citep{simonyan2014very}, and the multi-task layers are randomly initialized with a Gaussian distribution. TraceNet is trained with stochastic gradient descent(SGD) at a learning rate of $0.5 \times 10^{-4}$ and a batch size of 1. During training, we find a large batch size will ignore the contribution to gradient descent from classes that have less samples, and a small value can alleviate the problem and decrease the impact of data imbalance. The training is conducted in 3 times, each for 150K iterations.

\subsubsection{Testing inference}
\begin{figure}[!h]
	\centering
	\includegraphics[width=0.48\textwidth]{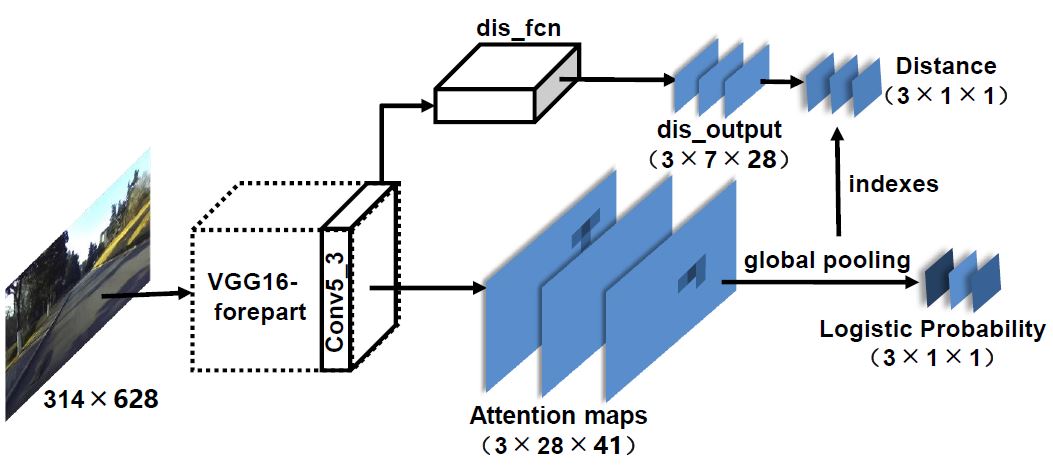}	
	\caption{Network structure of TraceNet while testing.}	
	\label{test-net}
\end{figure}
The prediction of complete affordances is conducted on the original image in a sliding window manner to find all possible driving actions while testing, as shown in Fig. \ref{test-net}. The network layers have all adopted convolutional operations and can be directly used on the original image. As a result, the attention maps and the distance branch will generate enlarged output sizes of $28\times 41$ and $7\times28$ respectively, as indicated with \textit{Attention maps} and \textit{dis\_output} in the figure. Due to the global max-pooling layer after the attention maps, the probability output for multi-label classification can always get a single maximum value from each attention map, indicating the existence of individual classes. We take a threshold of 0.9 for positive class identification. As the attention map has a fixed scale transformation to the original image, the position of maximum response in the attention map then can be used to localize visual attention center as well as the distance regression value.
\section{Experimental results}\label{experimental result}
\begin{figure}[!h]
	\centering
	\includegraphics[width=0.4\textwidth]{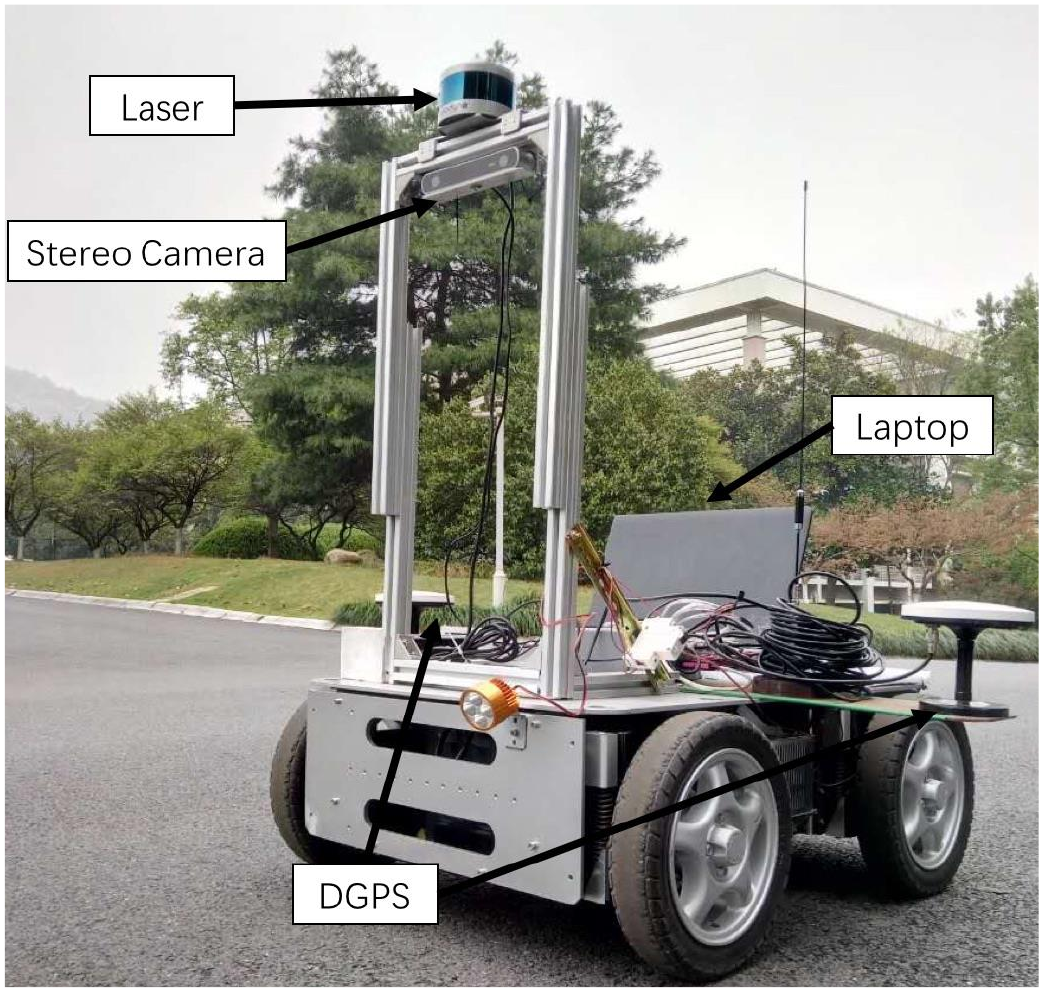}	
	\caption{Experiment vehicle.}	
	\label{vehicle}
\end{figure}

This section reports the experimental results of the proposed method, including the illustration of dataset, the result of trajectory segmentation, and the performance of affordances prediction. The proposed framework is validated on the datasets collected in our campus where most scenes do not have road lanes and traffic signs. The vehicle is a four-wheeled mobile robot equipped with a ZED stereo Camera and a Velodyne VLP-16 laser scanner, as shown in Fig .\ref{vehicle}. Only images from the left camera of ZED are used. The laser with 16 beams is used for global localization \citep{Tang2018From} which provided the vehicle trajectories. 

There are two datasets collected for training and test respectively:
\begin{itemize}
	\item \textbf{YQ21} is the observation of demonstration route (blue lines on campus map in Fig. \ref{outline}), which is recorded by driving a robot on a 1.2km route for 21 times. Each drive generates nearly 10000 images with a resolution of $314\times 628$ along with the vehicle poses.
	\item \textbf{YQ-South} is the observation of test route (red lines on campus map in Fig. \ref{outline}), which is collected by piloting robot on a 4.9km route, deploying unfamiliar surroundings for road affordances prediction test. This data has a total of 36000 images with a resolution of $314\times 628$. The ground truth is annotated by the projection of manually segmented trajectory.
\end{itemize}
Besides, the local region of YQ-South is utilized to test the network structure:
\begin{itemize}
	\item \textbf{YQ-South-Part} is the standard view region of YQ-South dataset. This setting is used to validate the network structure of TraceNet used for training(Fig. \ref{train-net}), as the Top1 branch and the distance branch can be simultaneously tested only with the standard view. This data has a total of 36000 images with a resolution of $224\times 224$.
\end{itemize} 

\subsection{Result of Trajectory Inference}
\begin{figure*}[!t]
	\centering
	\subfloat[3-1]{\includegraphics[width=0.19\textwidth]{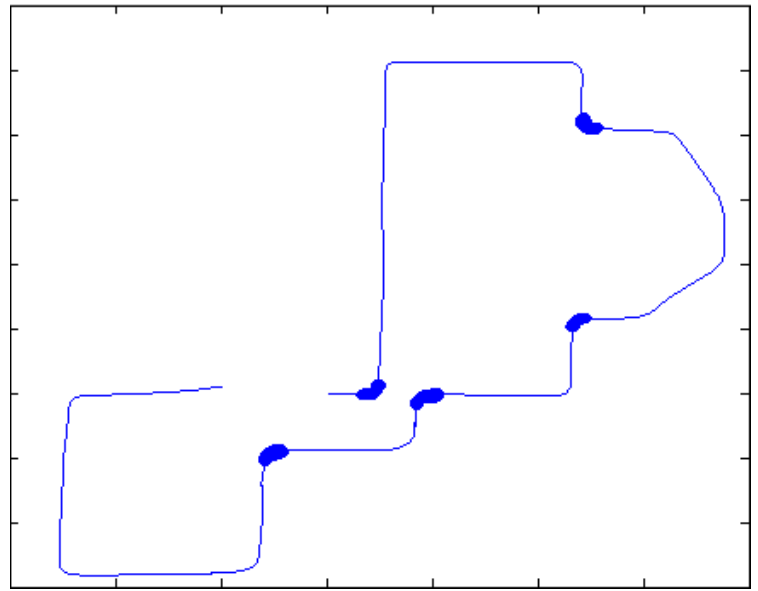}}	\subfloat[3-2]{\includegraphics[width=0.19\textwidth]{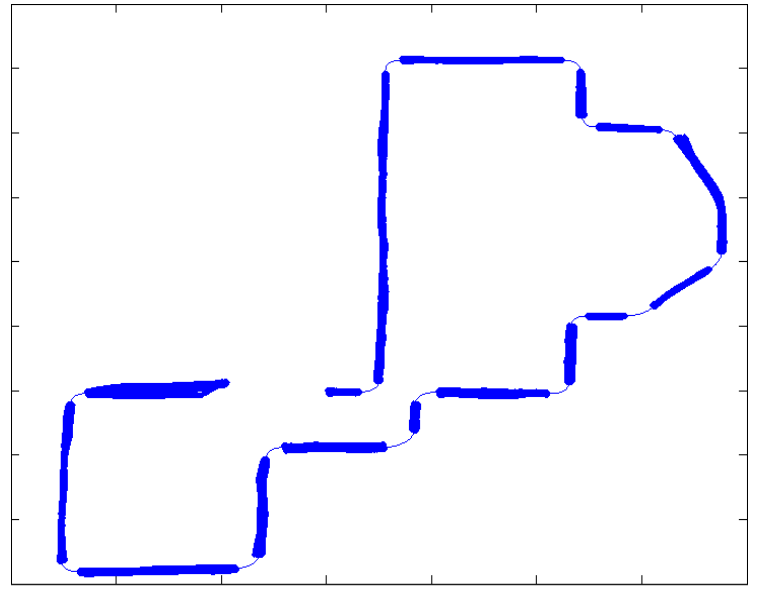}}
	\subfloat[3-3]{\includegraphics[width=0.19\textwidth]{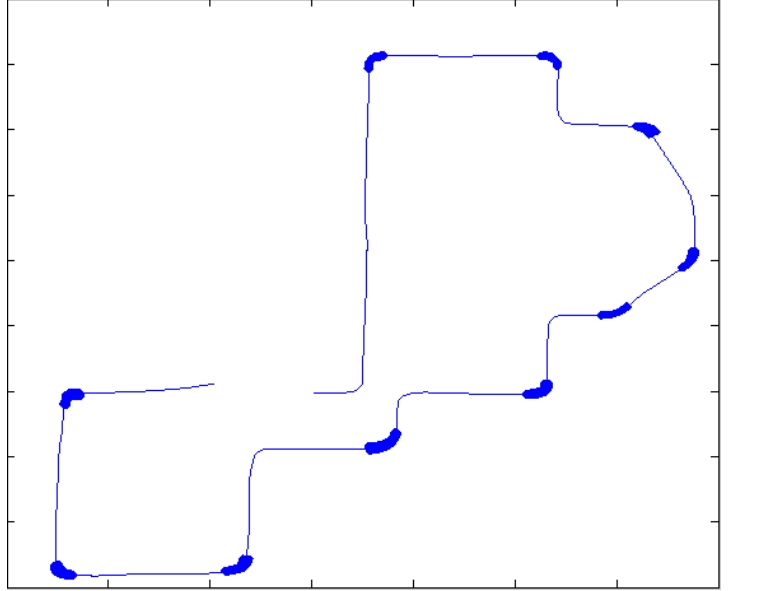}} 	\subfloat[5-1]{\includegraphics[width=0.19\textwidth]{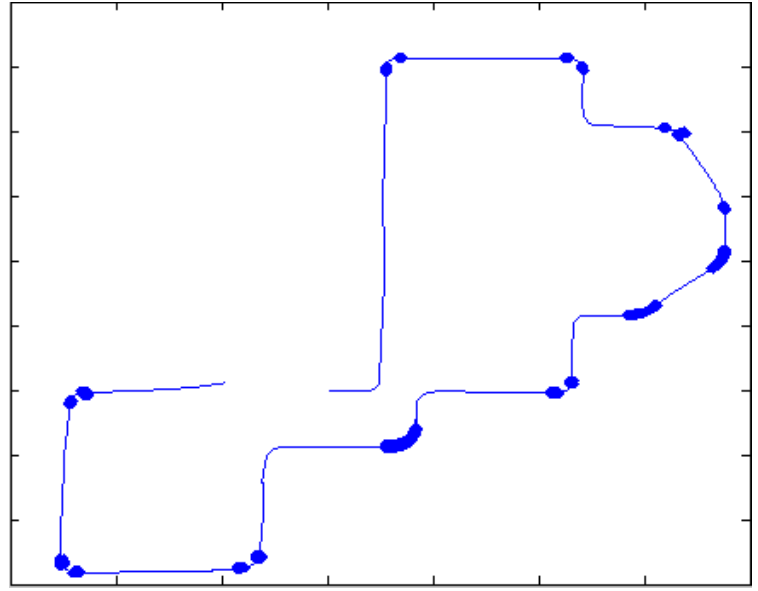}}	\subfloat[5-4]{\includegraphics[width=0.19\textwidth]{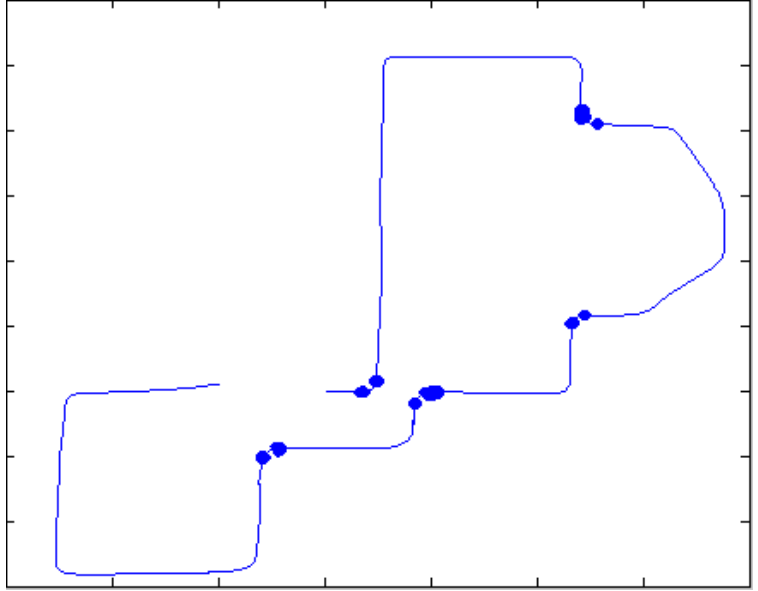}}
	\caption{Clustering results of 3-class actions and 5-class actions. (a), (b) and (c) are the 3-class result; (d) and (e) are the two more classes generated in the 5-class result. Blue single line shows the trajectory and bold line shows the clustering result of a certain class.}
	\label{5class}	
\end{figure*}

This part shows the result of trajectory segmentation for driving action analysis. We utilize the implementation of sticky HDP-HMM model\footnote[1]{https://homes.cs.washington.edu/~ebfox/software/} to cluster the sequential angular speeds. By tuning the truncation level of Dirichlet process and Gaussian emission model, different number of action classes can be obtained on the demonstrated trajectory, among which the results of 3-class and 5-class are intuitively meaningful as shown in Fig. \ref{5class}.
\begin{figure*}[h]
	\centering
	\includegraphics[width=0.95\textwidth]{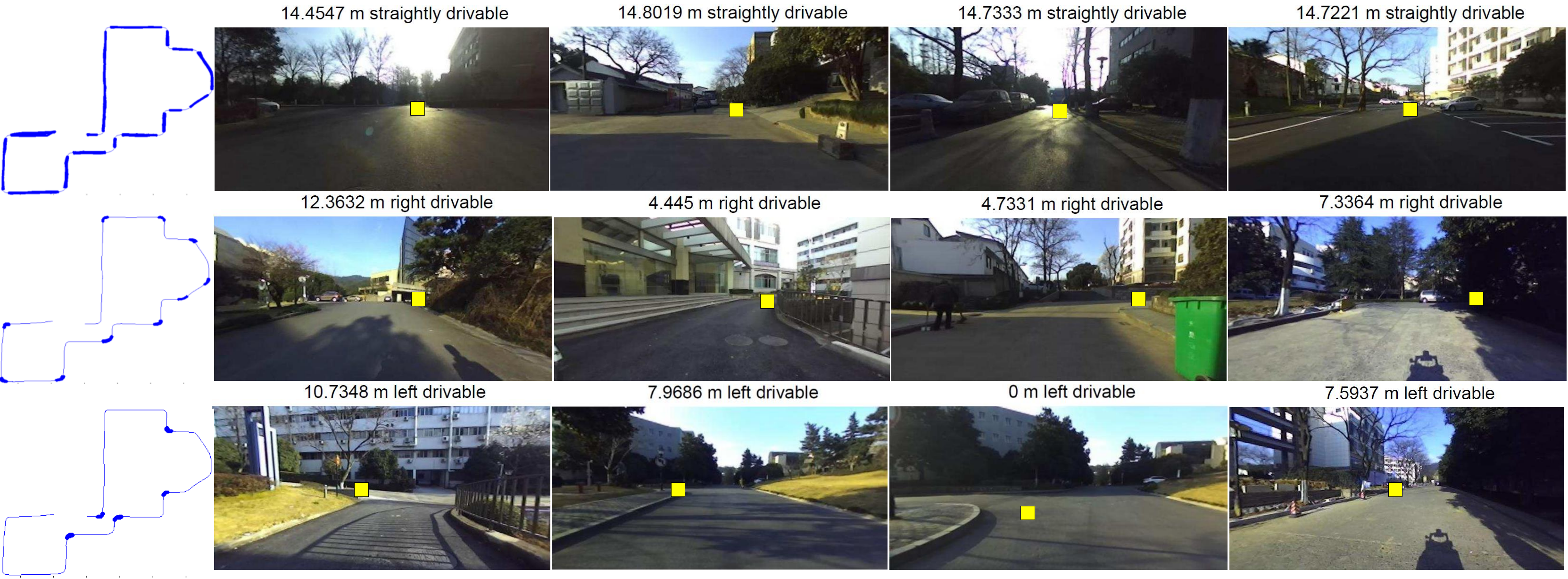}	
	\caption{Samples of partially labeled training data. From top to bottom： \textit{straight-drivable}, \textit{right-drivable} and \textit{left-drivable}. The yellow squares show the annotated attention center, which are enlarged for visualization.}
	\label{training data}
\end{figure*}

Fig.\ref{5class}(a)$\sim$(c) show the 3-class segmentation results. The order of classes are determined by the model and the driving actions can be subjectively assigned as the three primitive driving patterns of \textit{left-drivable}, \textit{straight-drivable} and \textit{right-drivable}. In the 5-class segmentation result, there are two more classes introduced as \textit{weakly-right/left-drivable} besides the 3-class result, as shown in Fig.\ref{5class}(d)$\sim$(e), which approximately represents the intervals of turn-in and turn-out areas of each road curve. To continue increasing the truncation levels, more classes can be generated but with less visually distinctive semantics. We consider more driving features can be involved in this model for driving action analysis in the future work, e.g. the recorded velocity can be used to assign the places where constant stops are needed for crowdy traffic.

In this paper, we primarily focus on the basic 3-class results for the visual affordances prediction. The road affordances of demonstrated driving action can then be annotated on image as illustrated in Section \ref{annotation}. Some samples of the partial annotation are shown in Fig. \ref{training data}. The information of absent labels is currently unknown.

\subsection{Result of Affordances Prediction}
This section presents the affordances learning results of the proposed TraceNet. To begin with, the network structure of TraceNet is verified based on the test results of YQ-South-Part. Then the road affordances prediction results are provided on both YQ21 and YQ-South.

\paragraph{Evaluation Metrics.} The performance of multi-label classification is measured on both the proportion of correctly judged drivable directions and the proportion of completely predicted images. Besides, a further analysis of precision, recall as well as binary classification accuracy for each single turning class are also provided. The accuracy of attention center localization is measured with the horizontal and vertical grid offsets (only the demonstrated driving action is considered for the test data). And distance regression is measured with an overall L1 error which is also calculated for each individual class. As for the Top1 branch, it is evaluated with the multi-class classification accuracy along with the confusion matrix.
\subsubsection{Validation of Network Structure}
\begin{figure}[!h]
	\centering
	\includegraphics[width=0.49\textwidth]{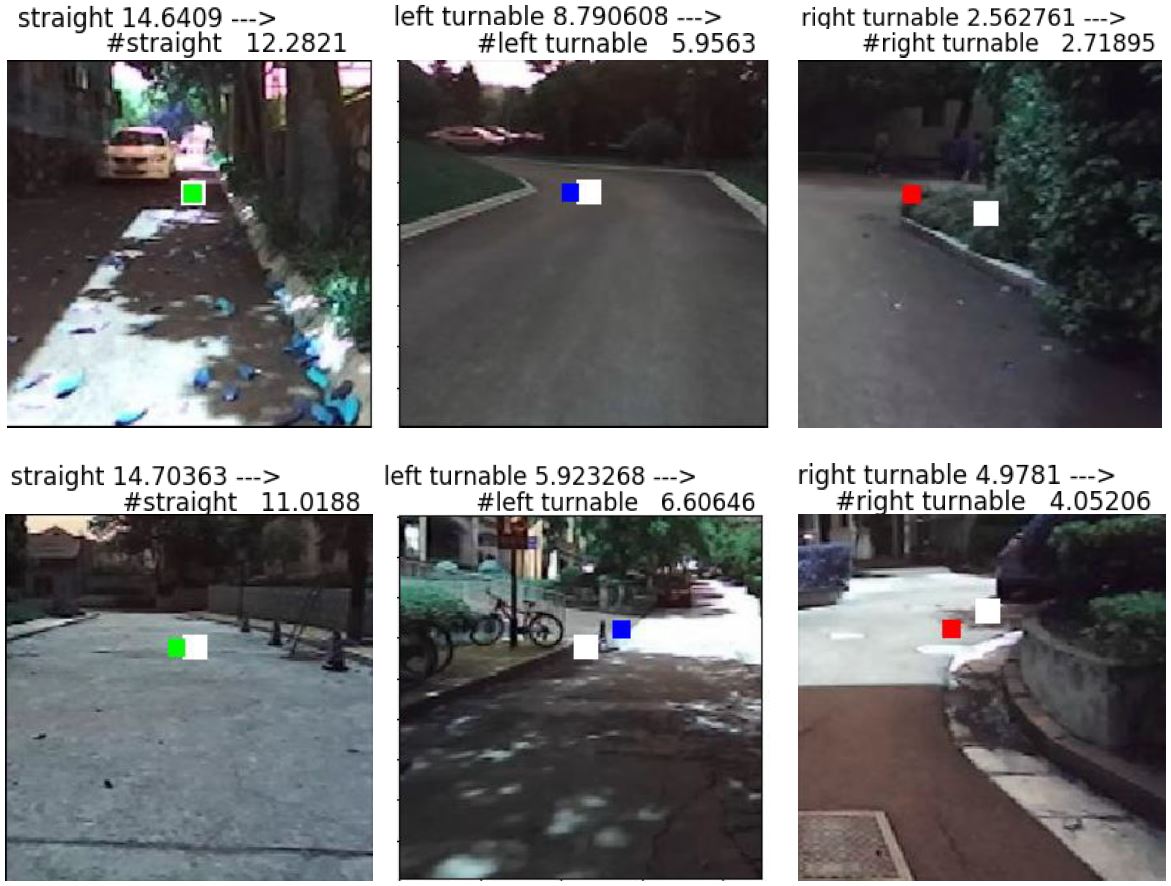}	
	\caption{Visual results of road affordances prediction on YQ-South-part.  The text on the top of image gives the ground truth and the predictions of allowed drivable directions and their remaining distances. White squares represent the ground truth attention centers and the colored squares denote the predictions. The ground truth is obtained by inference from manually segmented trajectories, thus some of the attention labels may be less accurate.}
	\label{south-part}
\end{figure}

YQ-South-Part is the standard view of YQ-South dataset, which only contains one drivable direction. It covers a longer route and can be used to validate network performance. Some visual results of affordances prediction are provide in Fig. \ref{south-part}. 

\begin{table*}[b]
	\caption{N{\scriptsize ETWORK} P{\scriptsize ERFORMANCE} {\scriptsize ON} YQ-SOUTH-PART }
	\centering
	\begin{tabular}{lccccccccc}
		\toprule[0.025cm]
		&\multicolumn{3}{c}{Top1\%}&\multicolumn{3}{c}{ Multi-label\%}&\multicolumn{2}{c}{ Attention (u/v)} & \multirow{2}{*}{Dis/m} \\			 
		\cmidrule(lr){2-4} \cmidrule(lr){5-7}  \cmidrule(lr){8-9}  
		& s &r & l& p & r&acc & u&v&
		\\
		\cmidrule{1-10}
		\textit{s(gt)} & \textbf{ 94.9}&4.0&1.0&93.6&92.6 &89.6 &0.81 &0.10 &3.24 \\
		\cmidrule{1-10}
		\textit{r(gt)}&20.4&\textbf{79.6}&0.0&66.5 &71.8&94.3  &2.30  &0.54 &3.85\\
		\cmidrule{1-10}
		\textit{l(gt)}&27.9&3.8&\textbf{68.3} & 92.4&59.6&93.0  &1.51 &0.50 &3.85\\
		\cmidrule{1-10}
		\textit{Acc}& 	\multicolumn{3}{c}{89.4}  & \multicolumn{3}{c}{85.4}   & 0.99    &0.17 &3.35\\  
		\bottomrule[0.025cm]
		\multicolumn{10}{l}{{\scriptsize s/r/l: straight/right/left drivable; p/r: precision/recall; u/v: horizontal/vertical grid offset;}}
	\end{tabular}
	\label{yqsouth-part}
\end{table*}
The overall quantitative evaluation is presented in Table \ref{yqsouth-part}. As can be seen, Top1 gets an accuracy of 89.4\% which is the 3-class classification performance. \textit{Right}/\textit{Left} classes are less accurate than that of \textit{straight}, which is mainly attributed to the data imbalance as the samples in straight class are much more than that in turning class. Moreover, the semantic ambiguity around the distance threshold to separate straight class with turning classes also has a more severe impact on the turning classes. Nevertheless, there is little confusion between \textit{right} and \textit{left} and the error rates are 3.8\% and 0\% respectively.

Considering multi-label task needs to address the absence/existence of each drivable class, it is actually a 8-class classification problem and a slight drop of 4\% accuracy is observed compared with the Top1 performance. For the individual classes, precision and recall for turning classes are generally less than the \textit{straight} class, while the accuracies have achieved higher value. This is mainly due to the presence of true negatives while calculating accuracies, as the massive correct predictions of straight-only image are the true negatives for turning classes. Besides, the result shows a big difference on precision and recall of the two turning classes. The insights on this can be found in the confusion matrices provided in Table \ref{binary-confusion}.

\begin{table}[h]
	\caption{M{\scriptsize ULTI-LABEL} C{\scriptsize ONFUSION} M{\scriptsize ATRICES}}
	\centering
	\begin{tabular}{llcccccccc}
		\toprule[0.025cm]
		&\multicolumn{4}{c}{Recall\%}&&\multicolumn{4}{c}{ Precision\%}\\			 
		\cmidrule(lr){2-5} \cmidrule(lr){7-10} 
		&& s &r & l&&& s &r & l\\
		\cmidrule{1-5}\cmidrule{6-10}
		\multirow{3}{*}{$gt$}&\textit{s} &95.1&3.8&1.&\multirow{3}{*}{$pre$}&s&93.5&2.&4.5 \\
		\cmidrule{2-5}\cmidrule{7-10}
		&\textit{r}&19.4&80.6&0.&&r&\textbf{29.8}&66.5&3.6\\
		\cmidrule{2-5}\cmidrule{7-10}
		&\textit{l}&25.8&2.7&71.5&&l&7.6&0.&92.4\\		 
		\bottomrule[0.025cm]
		\multicolumn{10}{l}{{\scriptsize s/r/l: straight/right/left; gt: ground truth(default); pre: predictions;}}
	\end{tabular}
	\label{binary-confusion}
\end{table}

For the recalls, approximately one fifth of turning classes are misclassified as \textit{straight} class, which is similar with that of Top1 result. As for the precisions, we observed a different result for \textit{right} and \textit{left} classes. There are around 30\% \textit{right} predictions come from the \textit{straight} class, while which is only 7.6\% for the \textit{left} predictions. The reason lies in the right-side driving style during trajectory collection, which causes the vision observations for some \textit{straight} samples being rather similar to the \textit{right} class and decreases the inter-class variations.

The accuracy of attention center localization is measured by the position offset of predicted grid to the ground truth grid, which is around 1/0.2 for the standard views. The distance error is approximately 3.4m in regard to a maximum prediction of 15m which we consider can be further improved in the future work, as the distance regression is substantially affected by the multi-task co-training (this has been further analyzed in following report). Moreover, there is an inherent difficulty for distance regression with monocular vision.

Besides the overall performance, two comparison experiments are conducted to verify the multi-task loss function and the multi-view sampling in an ablation study manner.

\noindent\textbf{Multi-task loss function.} This part shows the benefit of training with multi-task loss function, which also reflects the information mining procedure with the limited trajectory projection. In this experiment, the three branches of sub-tasks are gradually included for network training, and the result is presented in Table \ref{loss}. 
\begin{table}[!h]
	\caption{ M{\scriptsize ULTI-TASK} S{\scriptsize TUDY} {\scriptsize ON} YQ-SOUTH-PART }
	\centering
	\begin{tabular}{lcccc}
		\toprule[0.025cm]
		\multicolumn{1}{l}{\multirow{2}{*}{}} &
		\multicolumn{4}{c}{ Accuracy }\\			 
		\cmidrule(lr){2-5}   
		tasks & T\% & M\% & Attention (u/v) & D/m
		\\
		\cmidrule{1-5}
		\textit{T} &  89.0  &  -  & - & -   \\
		\cmidrule{1-5}
		\textit{T+M}& 89.7  & 84.3  & 1.48/1.05 & -  \\
		\cmidrule{1-5}
		\textit{T+M+D}& 89.4  & 85.5  & 0.99/0.17 & 3.35   \\
		\bottomrule[0.025cm]
		\multicolumn{5}{l}{{\scriptsize T/M/D: Top1/Multi-label/Distance;}}
	\end{tabular}
	\label{loss}
\end{table}

The single branch of Top-1 achieves an accuracy of 89.0\% for 3-class classification. Then the inclusion of multi-label branch has slightly improved Top1 accuracy while provided more information of individual possibility on driving classes and their related driving attention center. With the inclusion of distance regression, Top1 accuracy stays at a similar value while both multi-label prediction and attention center localization get improved results. Although the distance information is relatively less precise due to the limited image resolution, its quantitative measurement still offers a supplementary constraint for the search of attention center. 

\noindent\textbf{Multi-view sampling.} This part validates the contribution of different types of local views to the three training branches. In this experiment, different combination of sampling views are used for network training and the results are shown in Table \ref{multi-view}. \{S,P,N\} represents samples of standard view, positive view and negative view respectively.

\begin{table}[!h]
	\caption{ V{\scriptsize IEW} S{\scriptsize AMPLES} S{\scriptsize TUDY} {\scriptsize ON} YQSOUTH-PART}
	\centering
	\begin{tabular}{ccccccc}
		\toprule[0.025cm]
		\multicolumn{3}{c}{ View type} &
		\multicolumn{4}{c}{ Accuracy }\\			 
		\cmidrule(lr){1-3}\cmidrule(lr){4-7}   
		S&P&N & T\% & M\% & Attention (u/v) & D/m
		\\
		\cmidrule{1-7}
		\checkmark && &  87.1  &  82.8 &3.51/1.40&\textbf{2.00}   \\
		\cmidrule{1-7}
		\checkmark &\checkmark&& 88.0&85.1&1.81/1.35 &2.13\\
		\cmidrule{1-7}
		\checkmark&&\checkmark&88.6 &{\color{red}\textbf{ 14.8} }& 1.56/1.59 & 3.28\\
		\cmidrule{1-7}
		&\checkmark&\checkmark&88.4&81.4&1.74/1.08 &6.63 \\
		\cmidrule{1-7}
		\checkmark&\checkmark&\checkmark& \textbf{89.4}  & \textbf{85.4}  &  \textbf{0.99/0.17} & 3.35  \\
		\bottomrule[0.025cm]
		\multicolumn{7}{l}{{\scriptsize S/P/N: standard/positive/negative views;}}
	\end{tabular}
	\label{multi-view}
\end{table}

Top-1 accuracies have shown a tendency of ablation and tend to achieve better performance when more views are included. For multi-label task, the single standard view gets an accuracy of 82.8\%. A further inclusion of positive views contributes to 2.3\% increase of accuracy while the inclusion of negative views results in a significant accuracy drop to 14.8\%. The main reason for the failure of \textit{S+N} combination is attributed to the big data variation of the negative samples. The quantitative proportion of the two views are basically 1:1. However, negative views are sampled in a larger visual region that contains more appearance variations while standard view is a fixed region for each image. This particularly has a serious impact for the \textit{straight-drivable} class which has a high consistency on standard views and most attention maps of this class do not have a response. As a result, the error of attention center under this setting is mainly estimated from \textit{right} and \textit{left} classes, which shows a different ratio of u/v from other settings.

As for the attention center localization, either positive views or negative views contribute significantly, as implied in the second and third lines of Table \ref{multi-view}, which decrease around half the error in horizontal direction. When training with only randomly sampled views of\textit{ P+N} in the fourth line, performance on vertical grid offset further gets promotion. However, the related result of distance regression is remarkedly decreased to 6.63m, as random views do not provide necessary distance supervision. Apparently, distance regression gives its best performance of 2 m when only adopting the standard views. To consider a balanced performance for the three prediction tasks, the combination of all the local views leads to the best result, with a slight sacrifice on the precision of distance regression. 

\subsubsection{Complete Road Affordances Prediction}
\begin{figure*}[!h]
	\centering
	\includegraphics[width=\textwidth]{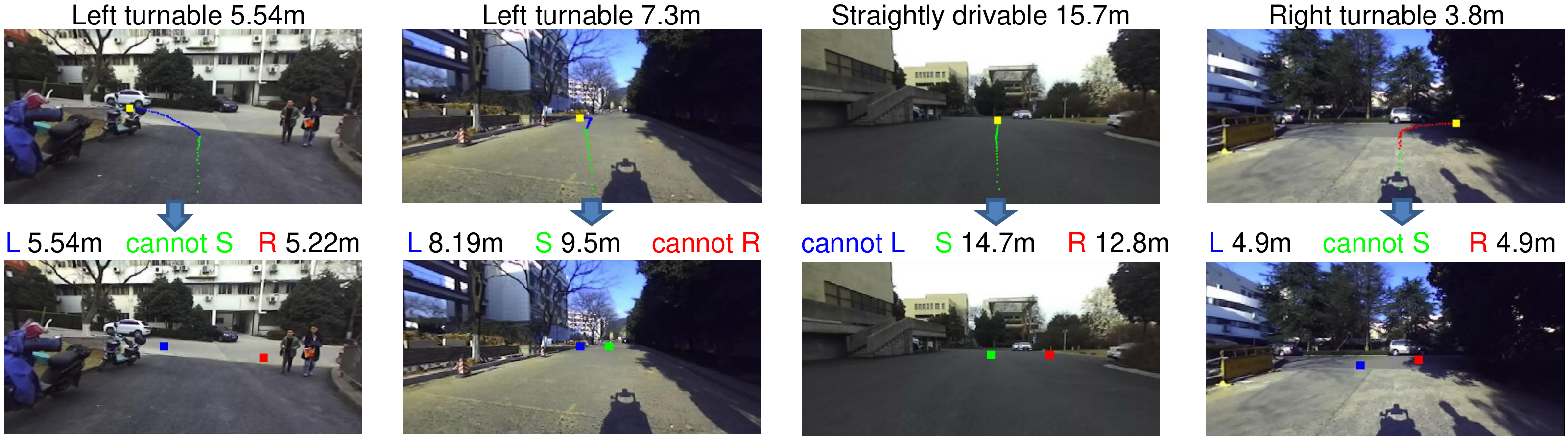}	
	\caption{Visual result of complete road affordances prediction on YQ21. The first row is the one-action affordances inferred from trajectory, and the second row is the complete affordances predicted in the TraceNet. L,S,R represent the three class of \textit{left, straight, right} respectively.}
	\label{yq21}
\end{figure*}

\noindent\textbf{Demonstration route of YQ21. }This part reports the prediction performance of TraceNet on the training route of YQ21, which complements the road affordances of other possible driving actions besides the demonstrated action. Fig. \ref{yq21} presents some visual results of completely predicted road affordances. The first row is the partial annotation inferred from trajectory and the second row is the complete affordances predicted from TraceNet. TraceNet runs at 25 fps for the inference on full image.

The quantitative result is presented in Table \ref{yq21-full}. There are two values indicating the accuracy of multi-label task as shown in the table, which are calculated based on the count of drivable directions and the count of images respectively. YQ21 achieves an overall accuracy of 88.2\% on directions and 74.3\% on complete images, with each single class reaching an accuracy of nearly 90\%. The recall rates of \textit{right} and \textit{left} are around 50\%, which implies much turning actions are missed and the positive training samples may not be sufficient in the partial annotation. The accuracies of attention center and distance regression are calculated on the correctly predicted classes, which is 1.53/0.52 grids offset and 0.79m respectively. 

\begin{table}[!h]
	\caption{P{\scriptsize REDICTION} P{\scriptsize ERFORMANCE} {\scriptsize ON} YQ21}
	\centering
	\begin{tabular}{lcccccc}
		\toprule[0.025cm]
		&\multicolumn{3}{c}{Multi-label\%}&\multicolumn{2}{c}{Attention} & \multirow{2}{*}{Distance/m} \\			 
		\cmidrule(lr){2-4} \cmidrule(lr){5-6}   
		& p &r & acc& u & v &
		\\
		\cmidrule{1-7}
		\textit{straight} &93.5&92.7&88.2&0.94 &0.46 &0.48 \\
		\cmidrule{1-7}
		\textit{right}&75.4&52.9&87.1&2.26 &0.63&2.38 \\
		\cmidrule{1-7}
		\textit{left}&76.9&44.5&89.0&2.45&0.56&1.82 \\ 
		\cmidrule{1-7}
		Accuracy& 	\multicolumn{3}{c}{88.2 / 74.3}  & 1.53&0.52   & 0.79\\  
		\bottomrule[0.025cm]
	\end{tabular}
	\label{yq21-full}
\end{table}

\noindent\textbf{Test route of YQ-South. } This part reports the affordances prediction result on the test route of YQ-South. The route of YQ-South covers that of YQ21, while a large proportion of the overlapped route drives in a reverse direction with totally different light conditions, thus contributing to a significant diversity in visual observations.
\begin{figure*}[ht]
	\centering
	\includegraphics[width=0.95\textwidth]{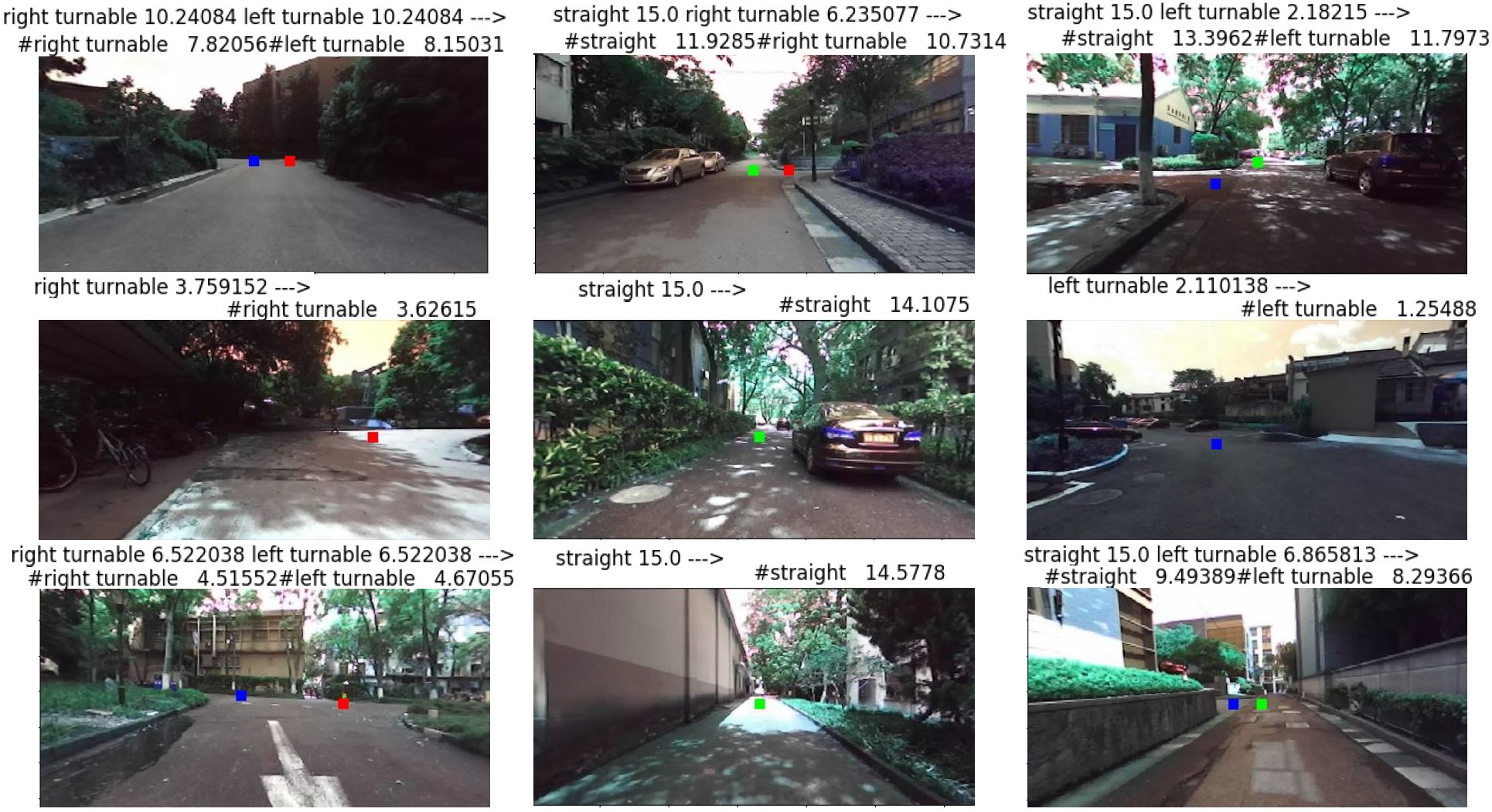}	
	\caption{Visual results of YQ-South. The text on the top of image gives the ground truth and the predictions of allowed drivable directions and their remaining distances. The attention centers are denoted with colored squares on the images. Green, red and blue represents \textit{straight}, \textit{right}, and \textit{left} respectively.}
	\label{south-full-1}
\end{figure*}
\begin{figure*}[!hb]
	\centering
	\includegraphics[width=0.9\textwidth]{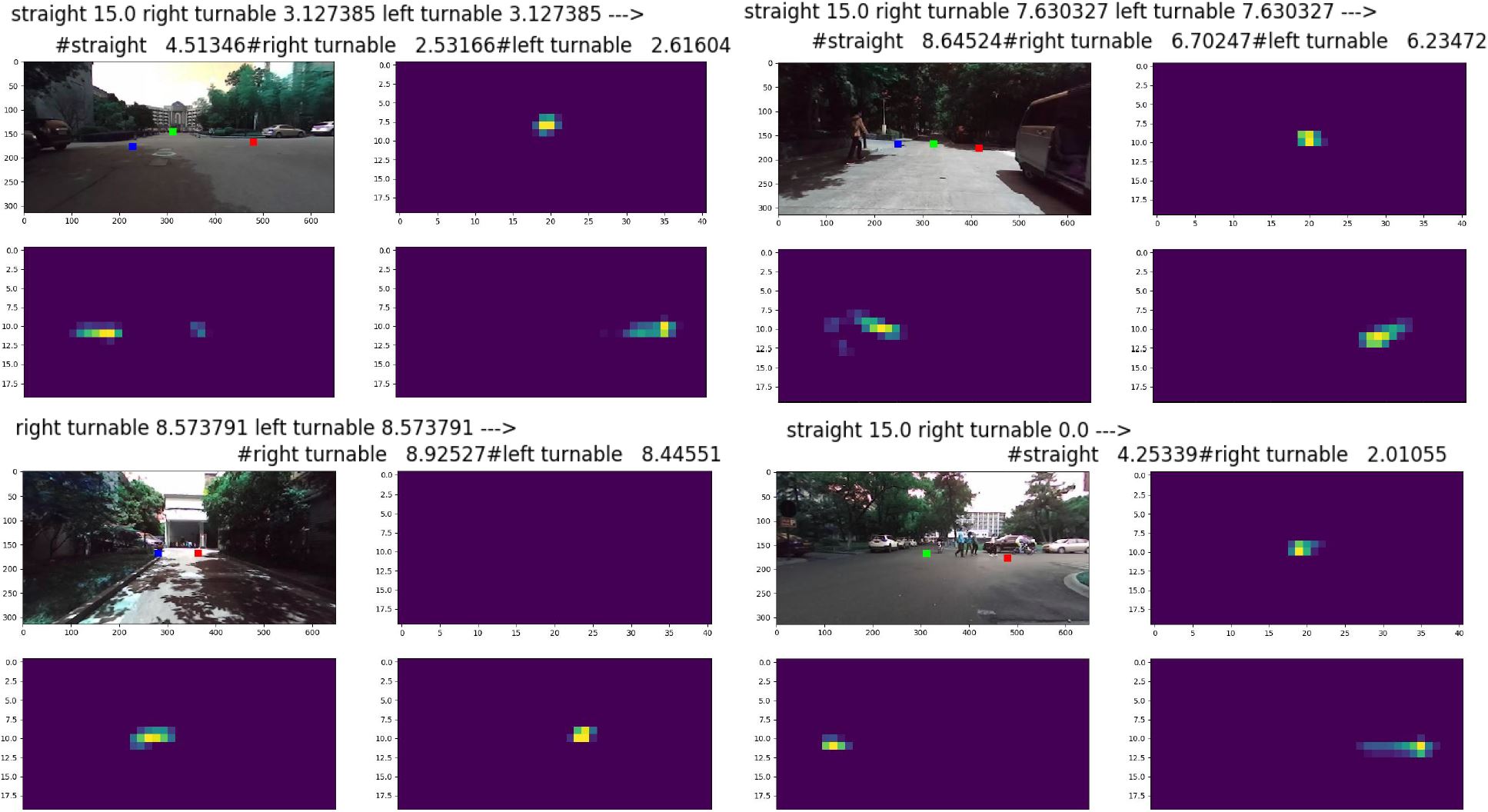}	
	\caption{Visualization of attention maps. The text on the top of each group gives ground truth and prediction results of drivable directions and their remaining distances. The colored squares in the top left figure of each group are the predicted attention centers. And the other three figures for each group are attention maps of straight drivable, left turnable and right turnable respectively.} 
	\label{south-full-2}
\end{figure*}

Fig. \ref{south-full-1} show some visual results on YQ-South. It can be inferred from the figure that distance regression has a large correlation among different classes, which tends to produce similar values when there are more than one positive classes in the image. This causes at least one large error on distance estimation especially for the turning classes, as shown in the last figure of the first row. The reason is partly attributed to the weak supervision from monocular vision and partly to the loss function of distance regression which does not have a negative suppression to false predictions. It can simultaneously encourage similar positive values for all the three classes. The correlation is more obvious between the turning classes, as samples from one turning class can be mirrored to the other class during the training. And the regression values become more relevant.

Fig. \ref{south-full-2} further provides the visualization of corresponding attention maps generated in the network. The operation of global max-pooling tries to converge the most distinctive features into one key point, thus the attention map of the complete image is rather clear with an explicit attention center.

The quantitative results are provided in Table \ref{yqsouth}. The overall accuracy of YQ-South is 80.9\% on directions and 66.2\% on images, and the accuracy for each single class is approximately 80\%. The performance is around 8\% less than that of YQ21. The accuracy of attention center prediction is 1.25/0.37 which has outperformed the demonstration route. As shown in the Table, this is mainly attributed to the significant number of correctly recognized straight-only road intervals, and the horizontal offset for turning classes is still large than the demonstration route. The distance regression result is 3.28m, and there is not much difference for different classes.

\begin{table}[!h]
	\caption{P{\scriptsize REDICTION} P{\scriptsize ERFORMANCE} {\scriptsize ON} YQ-South}
	\centering
	\begin{tabular}{lcccccc}
		\toprule[0.025cm]
		&\multicolumn{3}{c}{Multi-label\%}&\multicolumn{2}{c}{Attention} & \multirow{2}{*}{Distance/m} \\			 
		\cmidrule(lr){2-4} \cmidrule(lr){5-6}   
		& p &r & acc& u & v &
		\\
		\cmidrule{1-7}
		\textit{straight} &92.9&90.5&86.0&0.84&0.33 &3.246 \\
		\cmidrule{1-7}
		\textit{right}&62.2&39.0&81.6&3.92 &0.75&3.458 \\
		\cmidrule{1-7}
		\textit{left}&62.7&45.7&79.0&3.82&0.56&3.40 \\ 
		\cmidrule{1-7}
		\textit{Accuracy}& 	\multicolumn{3}{c}{80.9 / 66.2}  & 1.25 &0.37   & 3.28\\  
		\bottomrule[0.025cm]
	\end{tabular}
	\label{yqsouth}
\end{table}

\begin{figure}[!h]
	\centering
	\subfloat[Entrance of building.]{\includegraphics[width=0.238\textwidth]{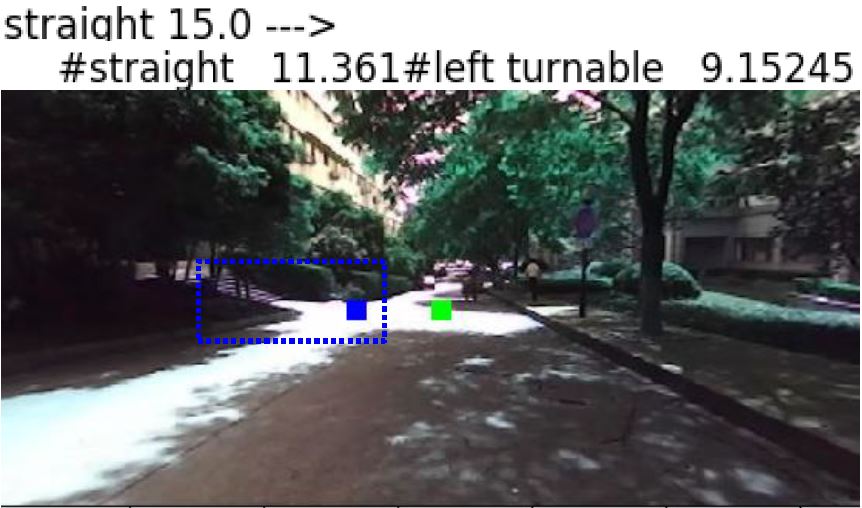}}
	\hspace{0.001\textwidth}
	\subfloat[Parking space.]{\includegraphics[width=0.235\textwidth]{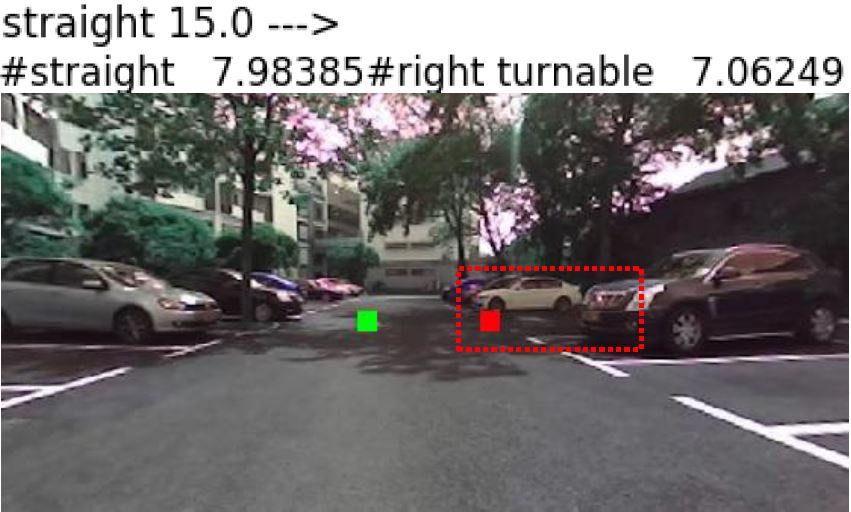}}
	\\	
	\subfloat[Occlusions]{\includegraphics[width=0.238\textwidth]{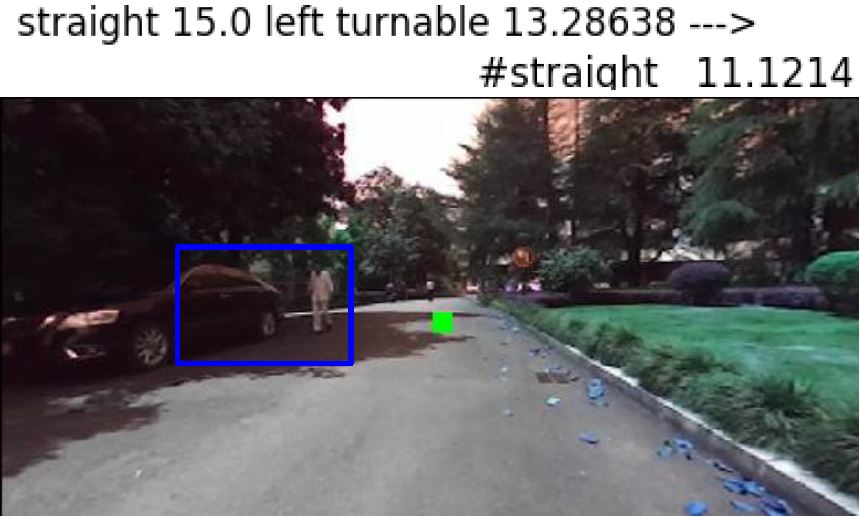}}
	\hspace{0.001\textwidth}
	\subfloat[Semantic ambiguity. ]{\includegraphics[width=0.235\textwidth]{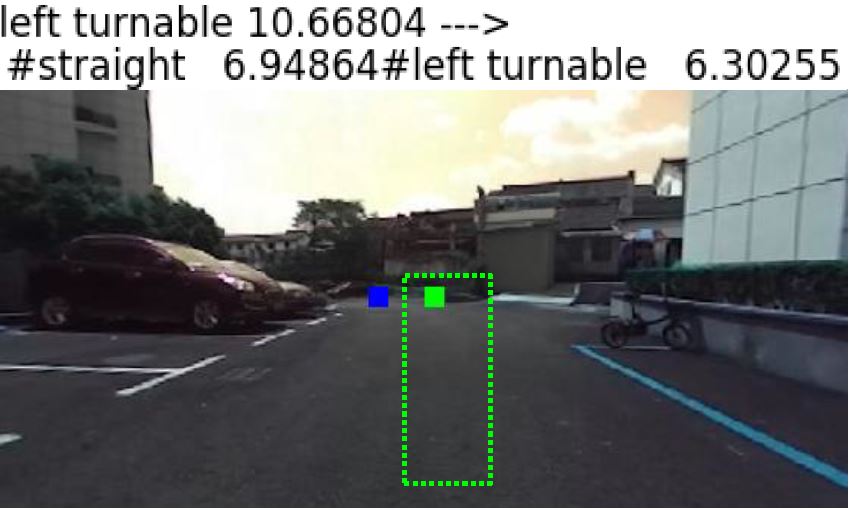}}
	\caption{Situations of incorrect prediction on YQ-South.}
	\label{errors}	
\end{figure}

\noindent\textbf{Challenges for prediction.} 
The reported performance on road affordances prediction has validated the proposes learning framework, while still needs further improvement for real applications. There are some typical challenges for this task as shown in Fig. \ref{errors}. The most probable false positives for \textit{right}/\textit{left} \textit{turnable} classes are the entrances of buildings and the parking spaces, where a possible turn may be allowed, as shown in Fig.\ref{errors}(a) and Fig.\ref{errors}(b). Since the robot has a limited field of view, it does not possess the knowledge behind turns. And we further did not model the differences of knowledge behind, which relates to more complicated cases. Besides, some highly occluded scenarios can lead to false negative predictions, as shown in Fig.\ref{errors}(c). The left turn is visually occluded by a car, causing a confusion even for human perceptions. And it is a common challenge in general visual perception tasks. Another kind of misclassification comes from semantic ambiguity around the distance threshold that separates straight class with turning classes, as shown in Fig.\ref{errors}(d), which is annotated with \textit{left-drivable} while still has a long area for straight driving. In general, variation relating to consecutive frames is a gradually changing process, and the representation of road affordances can be further improved to achieve more flexibility. Overall speaking, these challenges are also the critical problems faced with related researches on the scenes without traffic signs.

\section{CONCLUSIONS}\label{conclusion}
Road affordances understanding plays a major role for autonomous driving research, while most current works only focuses on the well-established urban road nets. This paper presented a direct inference and learning method for road affordances in the more complicated scenes without traffic signs. The method leverages vehicle trajectories to generate partial road affordances from driving behavior analysis, and further generalizes the deficient annotation to complete road affordances by introducing the unknown label for multi-task network training. We have specifically collected datasets for this recognition research and performed comprehensive experimental validation for each step. In the experiment, we started from vacant annotation on images and correctly parsed most of the road affordances under both familiar and unfamiliar surroundings. Our method does not need a predefined classification criterion and extra manual annotation on images, and hence it has the potential to produce more comprehensive traffic semantics with more driving behaviors performed and recorded. Our future work will consider to involve more driving features for trajectory analyses, and use the result for more comprehensive road affordances representation. 

\begin{acknowledgements}
	This work was supported in part by Science and Technology Project of Zhejiang Province (2019C01043) and in part by the National Nature Science Foundation of China (U1609210,61903332).
\end{acknowledgements}

\bibliographystyle{spbasic}      
\bibliography{mybib}   

\end{document}